%% file: article.tex
\pdfoutput=1

\documentclass{article}

\usepackage{times}
\usepackage{graphicx} 
\usepackage{subfigure} 

\usepackage{natbib}

\usepackage{algorithm}
\usepackage{algorithmic}

\newtheorem{theorem}{Theorem} 

\usepackage{amsmath,amssymb}

\usepackage{graphicx}
\usepackage{wrapfig,enumerate}
\usepackage{tikz}
\usetikzlibrary{decorations.markings}
\usetikzlibrary{positioning,shapes,backgrounds}
\usetikzlibrary{calc,arrows,intersections,patterns,fit}

\input{newcmds.tex}

\usepackage[nohyperref,accepted]{icml2016}

\icmltitlerunning{Node-By-Node Greedy Deep Learning for 
Interpretable Features}

\begin{document} 

\twocolumn[
\icmltitle{Node-By-Node Greedy Deep Learning for 
Interpretable Features
}

\icmlauthor{Ke Wu}{wuk3@rpi.edu}
\icmladdress{Department of Computer Science,
            110 8th Street, Troy, NY 12180 USA}
\icmlauthor{Malik Magdon-Ismail}{magdon@gmail.com}
\icmladdress{Department of Computer Science,
            110 8th Street, Troy, NY 12180 USA}

\icmlkeywords{greedy method, deep learning, node-by-node}

\vskip 0.3in
]
\input{absintro.tex}
\input{method.tex}

\input{expts.tex}

\bibliographystyle{icml2016}
\bibliography{mypapers,deep,deep2_kewu,dataset}

\end{document}

%% file: newcmds.tex
\def\math#1{$#1$}

\def\mand#1{$$#1$$}

\def\mandc#1{\mand{\abovedisplayskip=3.5pt plus 1pt minus 1pt%
\abovedisplayshortskip=0pt plus 1pt minus 1pt%
\belowdisplayskip=3.5pt plus 1pt minus 1pt%
\belowdisplayshortskip=0pt plus 1pt minus 1pt%
#1}}

\def\mldc#1{\mld{\abovedisplayskip=3.5pt plus 1pt minus 1pt%
\abovedisplayshortskip=0pt plus 1pt minus 1pt%
\belowdisplayskip=3.5pt plus 1pt minus 1pt%
\belowdisplayshortskip=0pt plus 1pt minus 1pt%
#1}}

\def\mld#1{\begin{equation}
#1
\end{equation}}

\def\R{\mathbb{R}}
\def\r#1{{(\ref{#1})}}

\newcommand{\mat}[1]{{{\mathrm{#1}}}}
\newcommand{\vect}[1]{{{\mathbf{#1}}}}

\def\norm#1{{\|#1\|}}

\newcommand{\xx}{\vect{x}}

\providecommand\remove[1]{}

\def\matW{\mat{W}}


%% file: absintro.tex
\begin{abstract} 
Multilayer networks have seen a resurgence under the umbrella
of deep learning. Current deep learning algorithms 
train the layers of the network
sequentially, improving algorithmic performance as well as 
providing some regularization.
We present a new training algorithm for deep networks which trains
\emph{each node in the network} sequentially. Our algorithm is
orders
of magnitude faster, creates more interpretable
internal representations at the node level, while 
not sacrificing on the ultimate out-of-sample performance.
\end{abstract} 

\section{Introduction}
Multilayer neural networks have gone through ups and downs since their
arrival in~\cite{rosenblatt1958,widrow1960,hoff1962}. The
resurgence in ``deep'' networks is largely due to the
efficient greedy layer by layer algorithms for training, that
create meaningful hierarchical representations of the data. 
Particularly in the era of ``big data'' from diverse 
applications, efficient training to create data representations that 
provide insight into 
the complex features captured by the neurons are important. We
explore these two dimensions of training a deep network.
Assume a standard machine learning from data setup \cite{malik173}, 
with \math{N} datapoints \math{(\xx_1,y_1),\ldots,(\xx_N,y_N)};
\math{\xx_n\in\R^{d}} and
\math{y_n\in\{0,1,\ldots,c-1\}} (multi-class setting).

\begin{wrapfigure}[10]{r}{0.2\columnwidth}
\vspace*{-0.225in}
\begin{center}
\hspace*{-0.2in}
\includegraphics[width=.2\columnwidth]{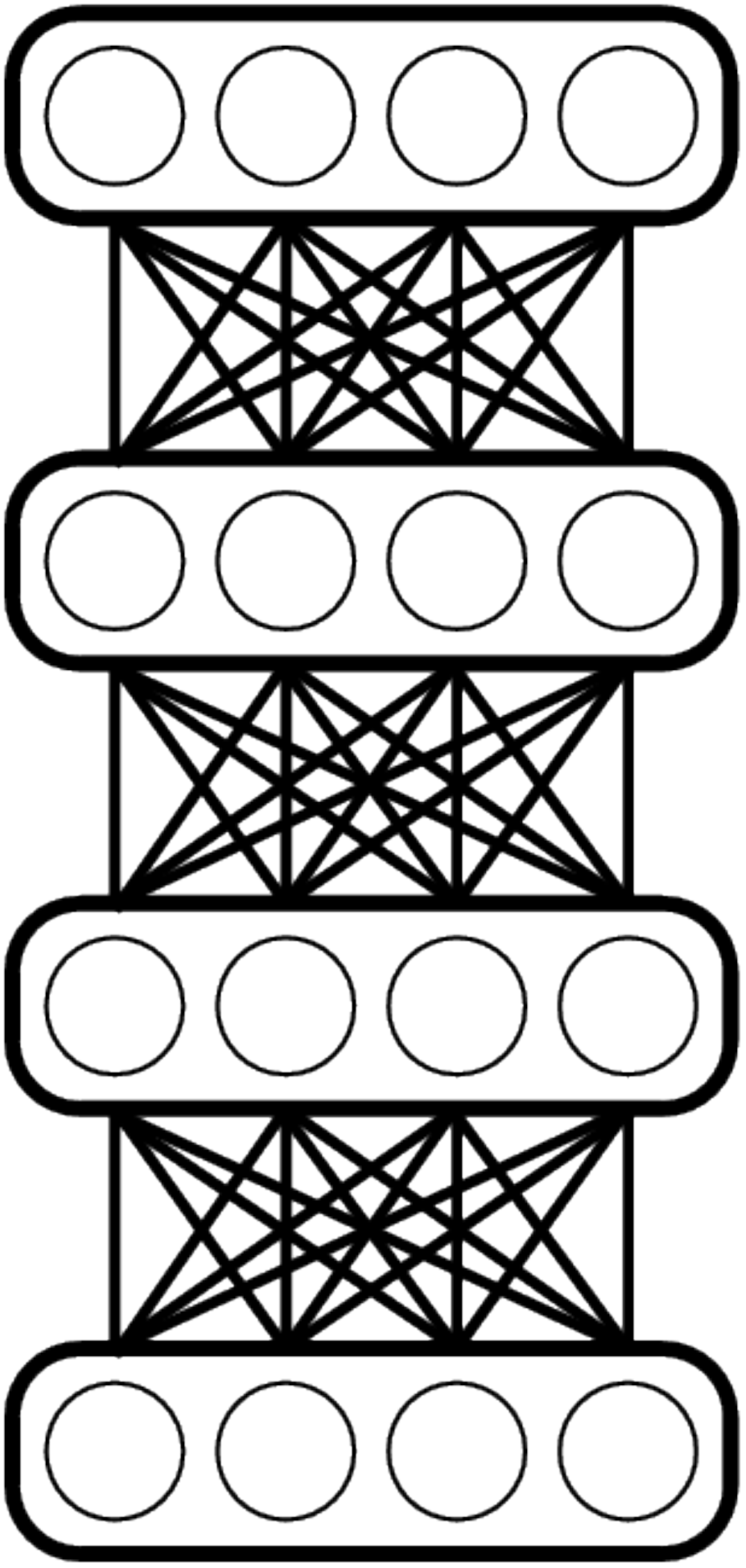}
\\[-2pt]
\hspace*{-0.2in}
\scalebox{0.8}{\bf Deep Network}
\end{center}
\end{wrapfigure}
We refer to~\citet[Chapter 7]{malik173} for 
 the basics of
multilayer networks, including notation which we very quickly 
summarize here. On the right, we show a feedforward network
architecture. Such a network is ``deep''
because it has many (\math{\gg 2}) layers.
We assume that a network architecture has been fixed.
The network implements a function whereby in each layer (\math{\ell}),
the output of the previous layer (\math{\ell-1}) is transformed into the
output of the layer \math{\ell} until one reaches the final layer on 
top, which is the output of the network. The function implemented
by 
layer \math{\ell} is 
\mandc{
\xx^{(\ell)}=\tanh(\matW^{(\ell)}\xx^{(\ell-1)}),
}
where \math{\xx^{(\ell)}} is the output of layer \math{\ell}, and 
the weight-matrix \math{\matW^{(\ell)}}
(of appropriate dimensions to map a
vector from layer \math{\ell-1} to a vector in \math{\ell}) are
parameters to be learned from data. The training phase
uses the data to identify all the parameters 
\math{\{\matW^{(1)},\matW^{(2)},\ldots,\matW^{(L)}\}}
of the deep network.

\begin{figure}[t]
\vskip 0.2in
\begin{center}
\centerline{
\begin{tabular}{c@{\hspace*{0.15in}}c@{\hspace*{0.15in}}c@{\hspace*{0.15in}}c}
\resizebox{0.15\columnwidth}{!}{\includegraphics*{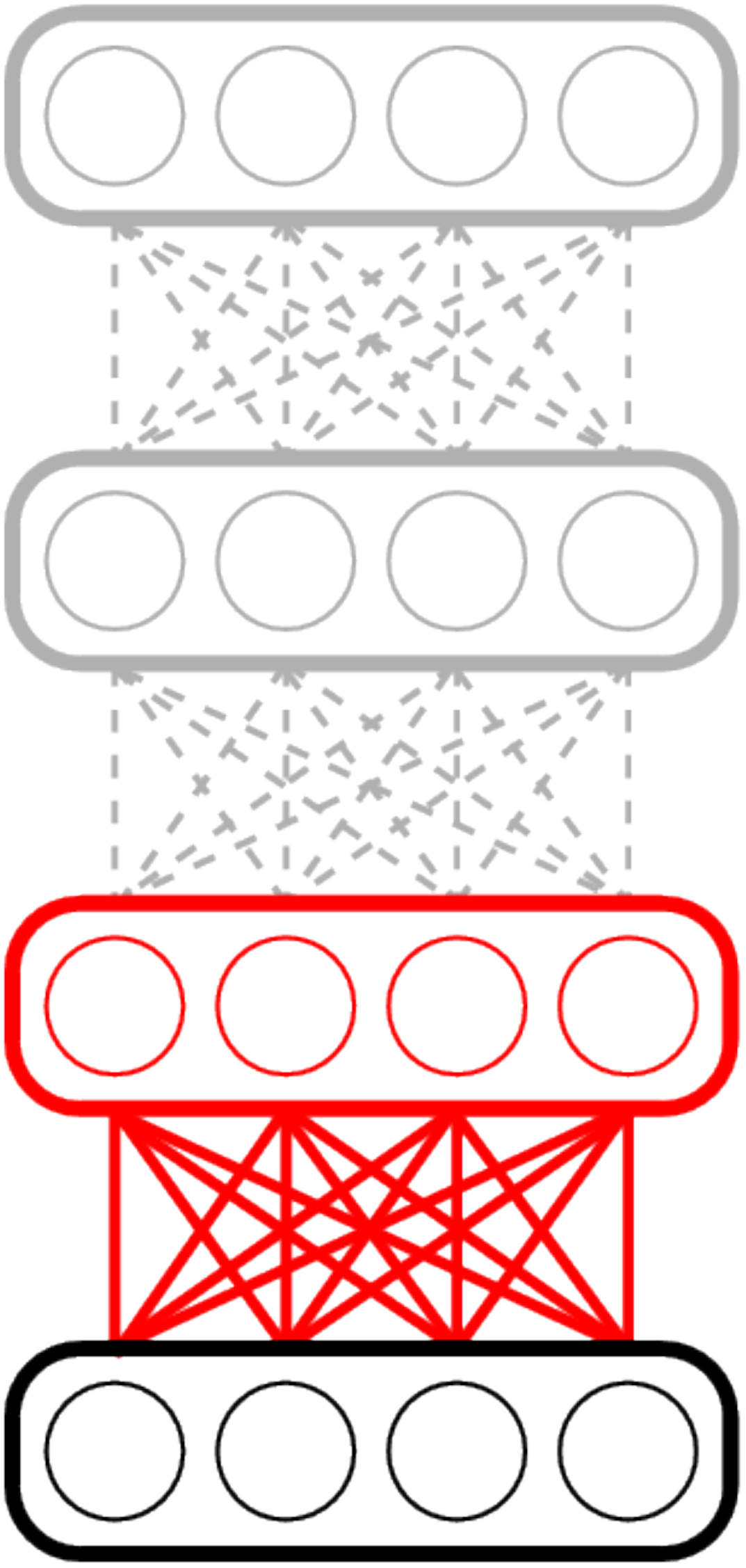}}&
\resizebox{0.15\columnwidth}{!}{\includegraphics*{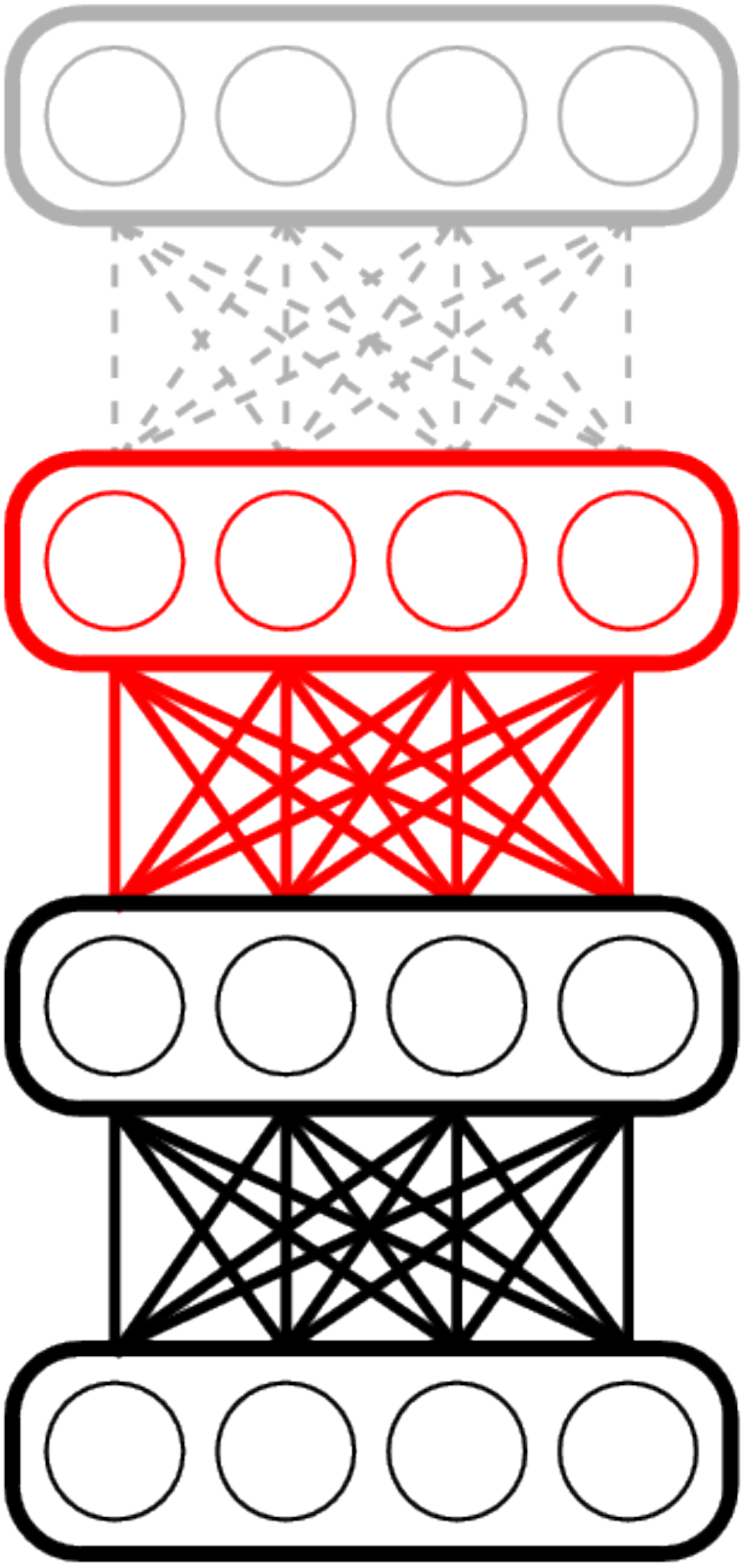}}&
\resizebox{0.15\columnwidth}{!}{\includegraphics*{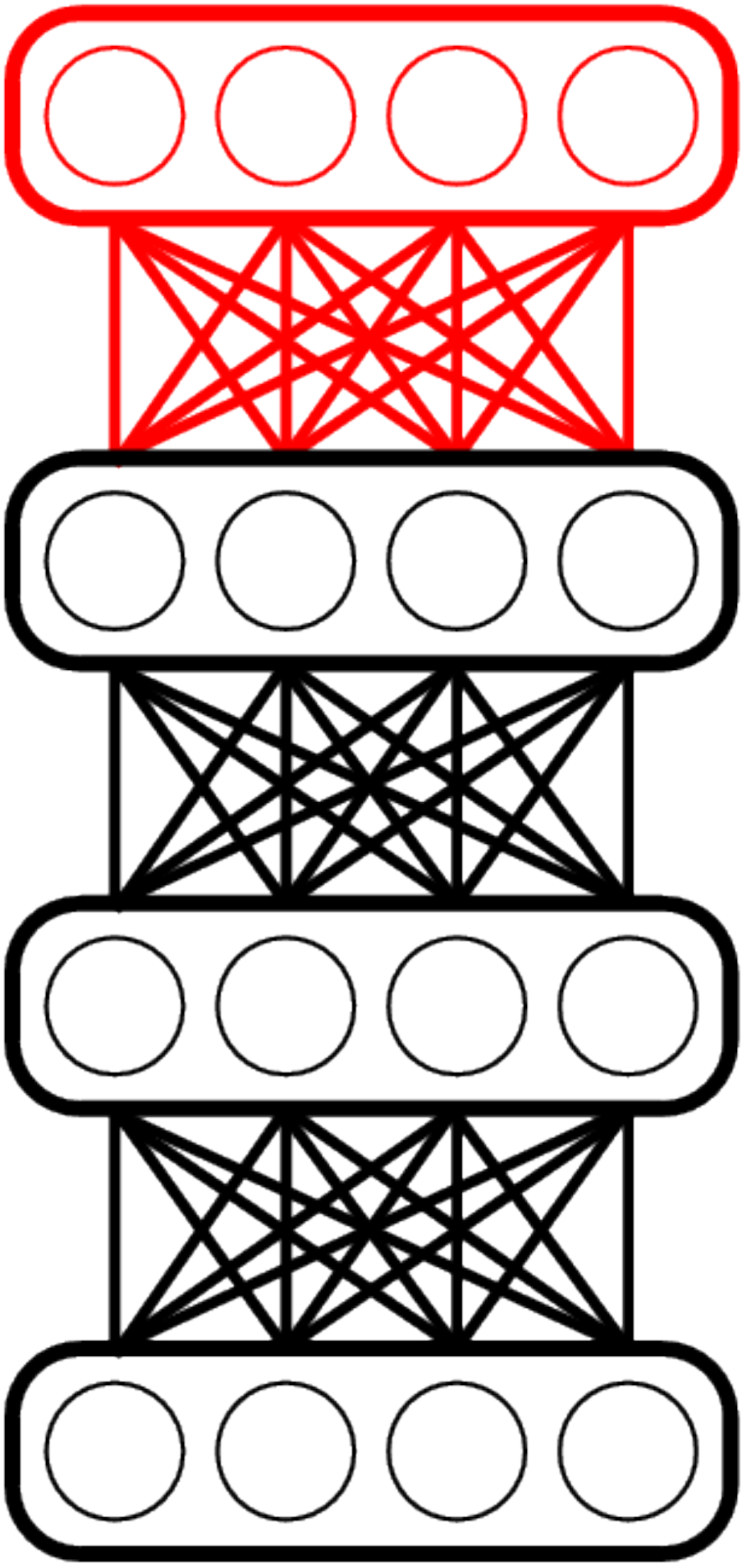}}&
\resizebox{0.15\columnwidth}{!}{\includegraphics*{DeepGreedy3.pdf}}
\\[2pt]
\small Learn \math{\matW^{(1)}}&
\small Learn \math{\matW^{(2)}}&
\small Learn \math{\matW^{(3)}}&
\small Fine tuning
\end{tabular}}
\caption{Layer-by-layer greedy deep learning algorithm.} 
\label{fig:layers}
\end{center}
\vskip -0.2in
\end{figure}

Backpropagation which trains all the weights simultaneously, 
allowing for maximum flexibility, 
was the popular approach to training a deep
network~\cite{rumelhart1986}.
The current approach is layer-by-layer:
train the first layer weights~\math{\matW^{(1)}};
then train the second layer weights~\math{\matW^{(2)}}, \emph{keeping the
first layer weights fixed}; and so on until all the weights are learned. 
In practice, once all the weights have been learned in the
greedy-layer-by-layer manner (often referred to as pre-training), 
the best results are obtained
by fine tuning all the weights using a few iterations of
backpropagation (Figure~\ref{fig:layers}).

\begin{figure*}[!ht]
\begin{center}
{\begin{tabular}{cc}
\setlength{\tabcolsep}{3pt}
\begin{tabular}{ccc}
\resizebox{0.3\columnwidth}{!}{\includegraphics[trim={4cm 0.5cm 4cm 0.5cm},clip] {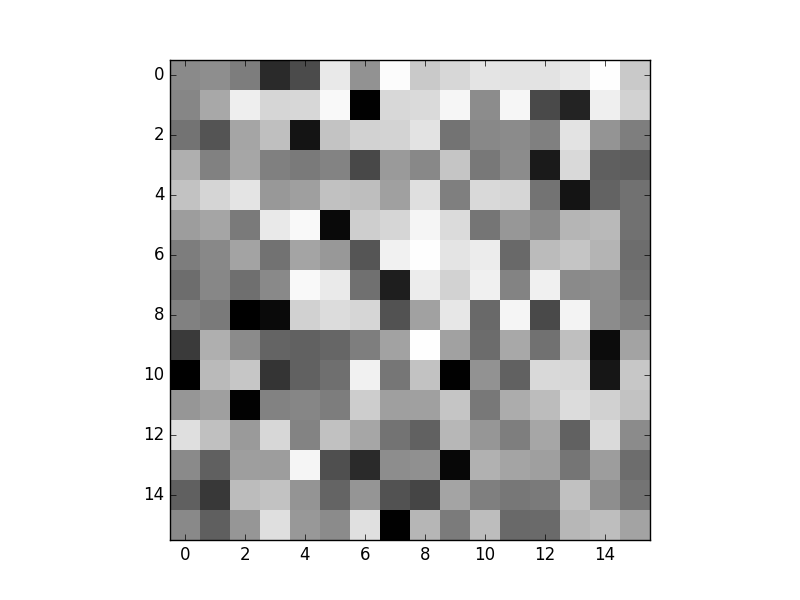}}&
\resizebox{0.3\columnwidth}{!}{\includegraphics[trim={4cm 0.5cm 4cm 0.5cm},clip]{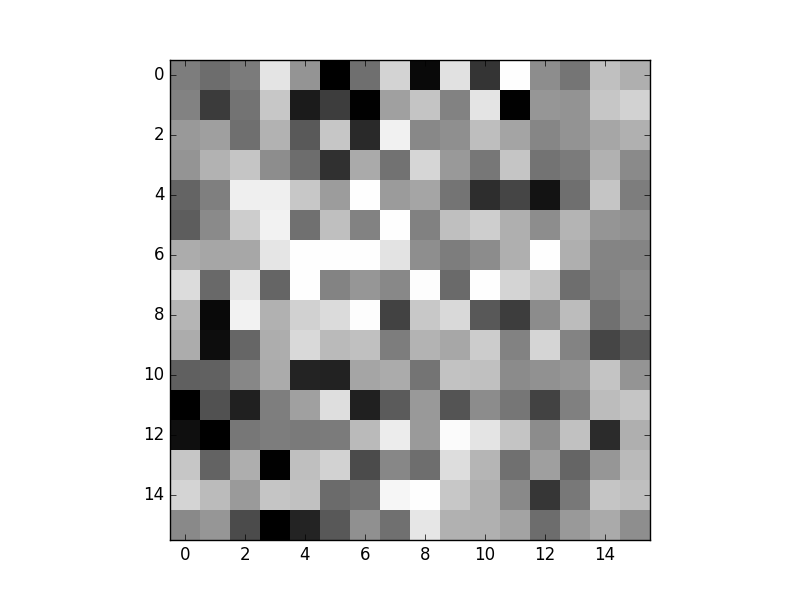}}&
\resizebox{0.3\columnwidth}{!}{\includegraphics[trim={4cm 0.5cm 4cm 0.5cm},clip]{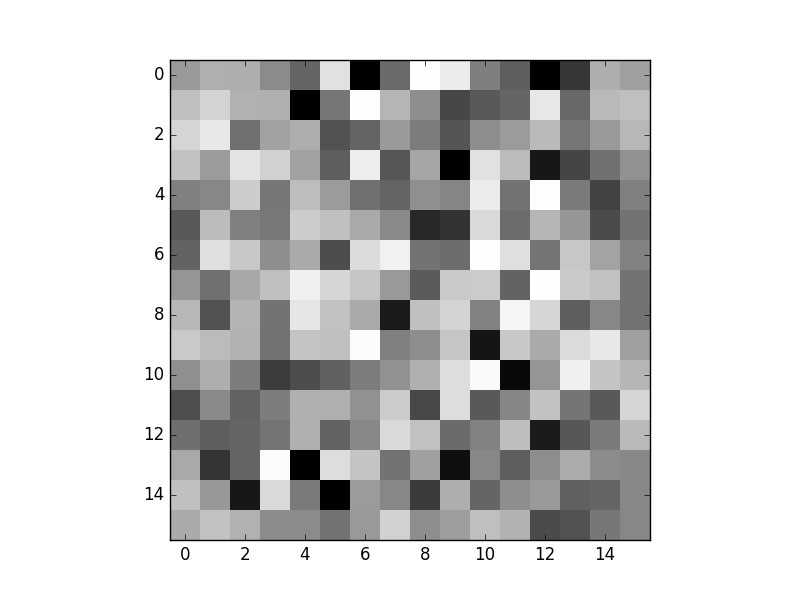}}\\
\resizebox{0.3\columnwidth}{!}{\includegraphics[trim={4cm 0.5cm 4cm 0.5cm},clip]{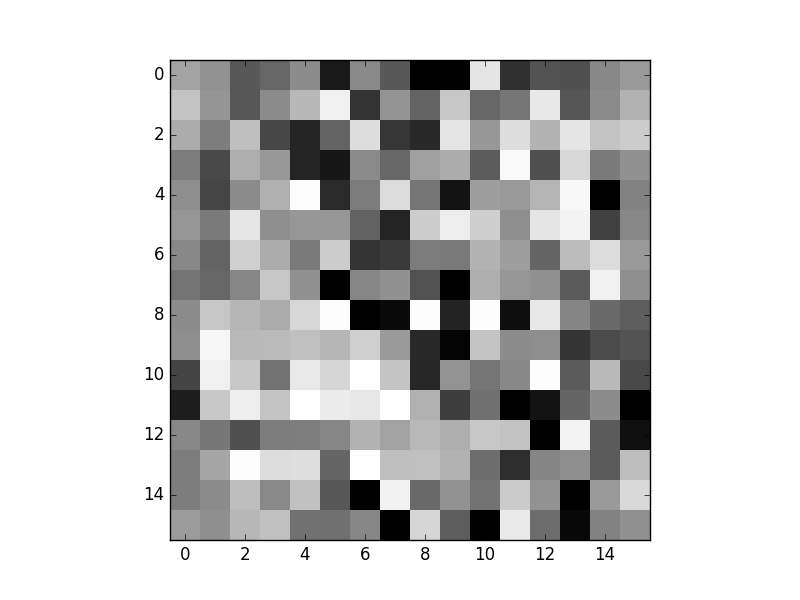}}&
\resizebox{0.3\columnwidth}{!}{\includegraphics[trim={4cm 0.5cm 4cm 0.5cm},clip]{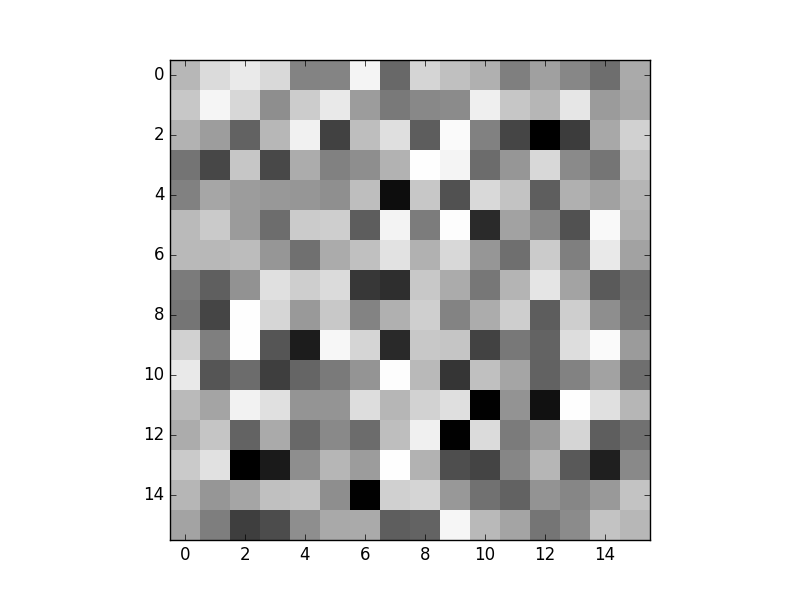}}&
\resizebox{0.3\columnwidth}{!}{\includegraphics[trim={4cm 0.5cm 4cm 0.5cm},clip]{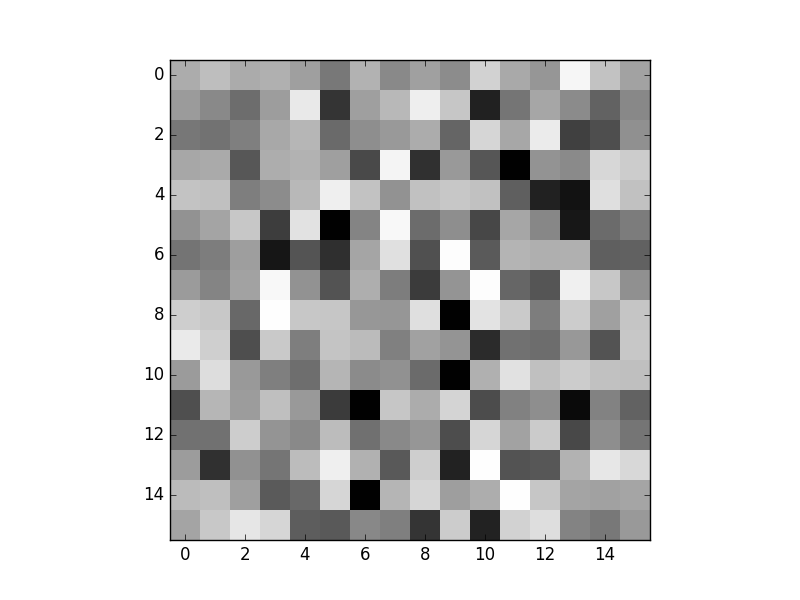}}\\
\resizebox{0.3\columnwidth}{!}{\includegraphics[trim={4cm 0.5cm 4cm 0.5cm},clip]{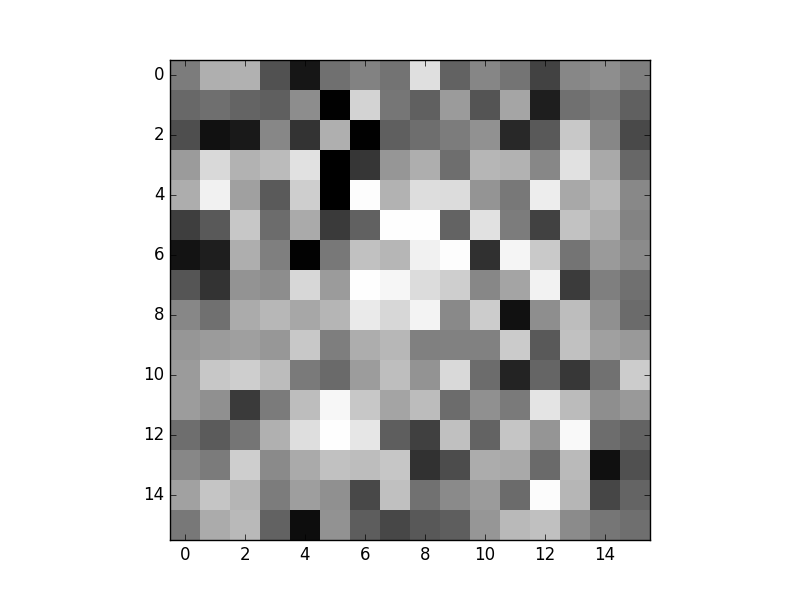}}&
\resizebox{0.3\columnwidth}{!}{\includegraphics[trim={4cm 0.5cm 4cm 0.5cm},clip]{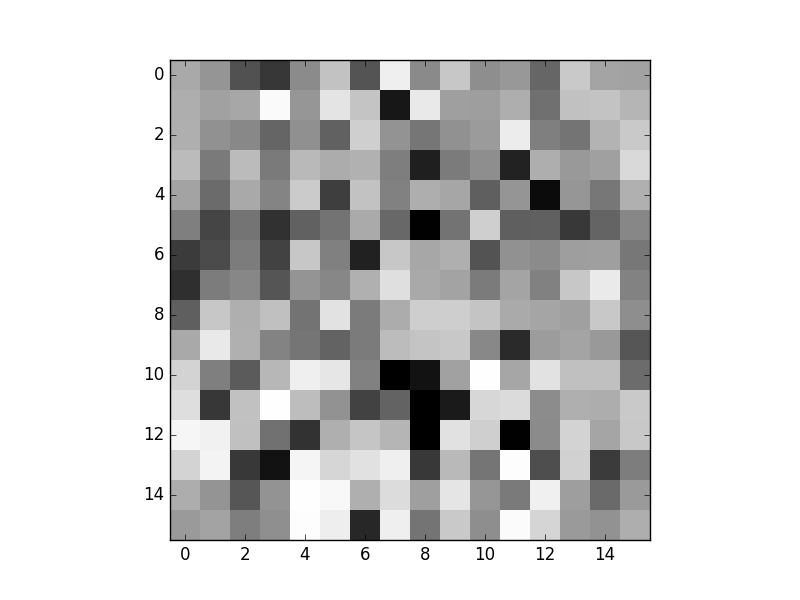}}&
\resizebox{0.3\columnwidth}{!}{\includegraphics[trim={4cm 0.5cm 4cm 0.5cm},clip]{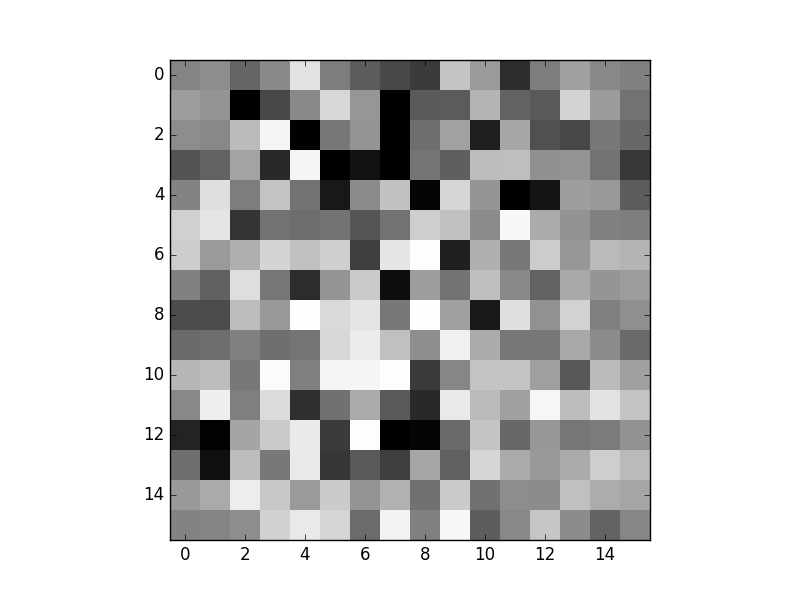}}\\
\end{tabular}
&
\setlength{\tabcolsep}{3pt}
\begin{tabular}{ccc}
\resizebox{0.3\columnwidth}{!}{\includegraphics[trim={4cm 0.5cm 4cm 0.5cm},clip]{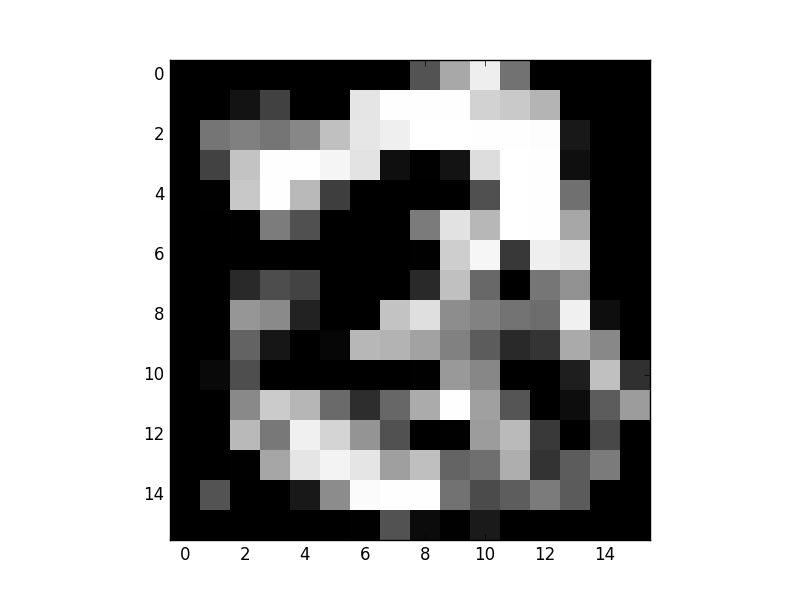}}&
\resizebox{0.3\columnwidth}{!}{\includegraphics[trim={4cm 0.5cm 4cm 0.5cm},clip]{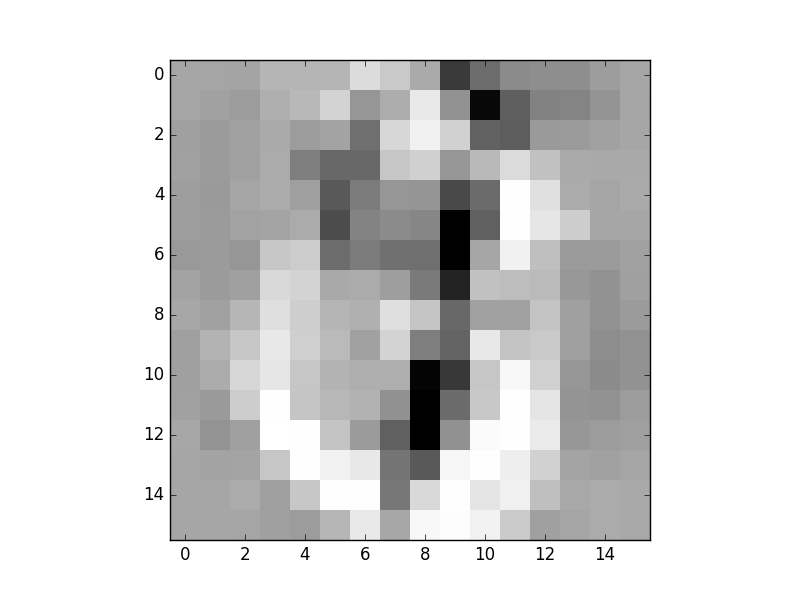}}&
\resizebox{0.3\columnwidth}{!}{\includegraphics[trim={4cm 0.5cm 4cm 0.5cm},clip]{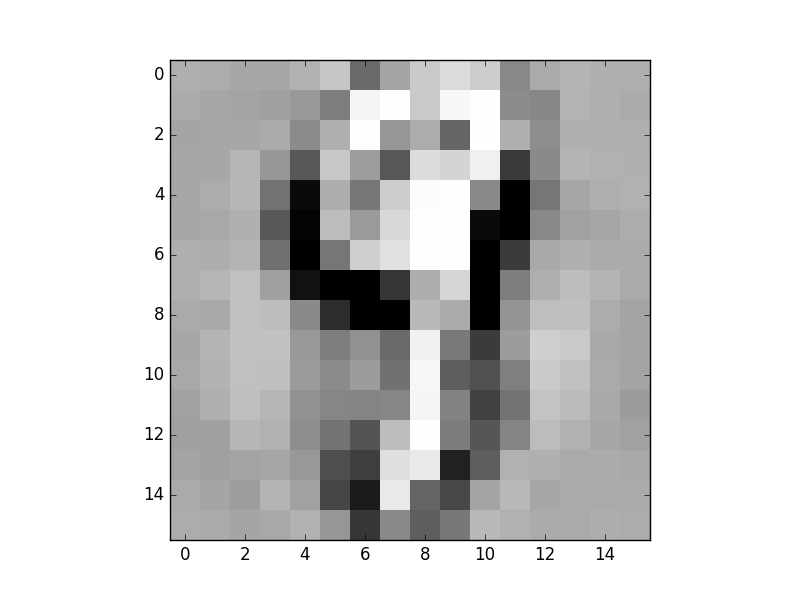}}\\
\resizebox{0.3\columnwidth}{!}{\includegraphics[trim={4cm 0.5cm 4cm 0.5cm},clip]{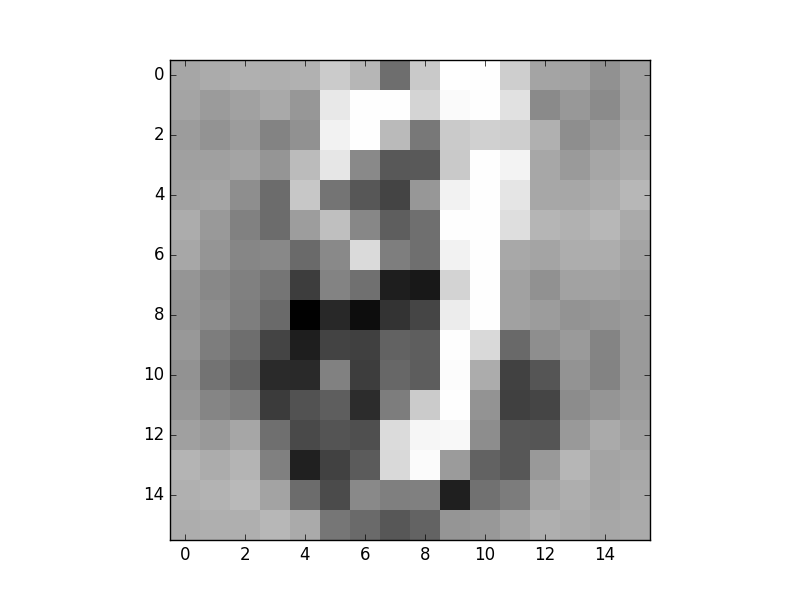}}&
\resizebox{0.3\columnwidth}{!}{\includegraphics[trim={4cm 0.5cm 4cm 0.5cm},clip]{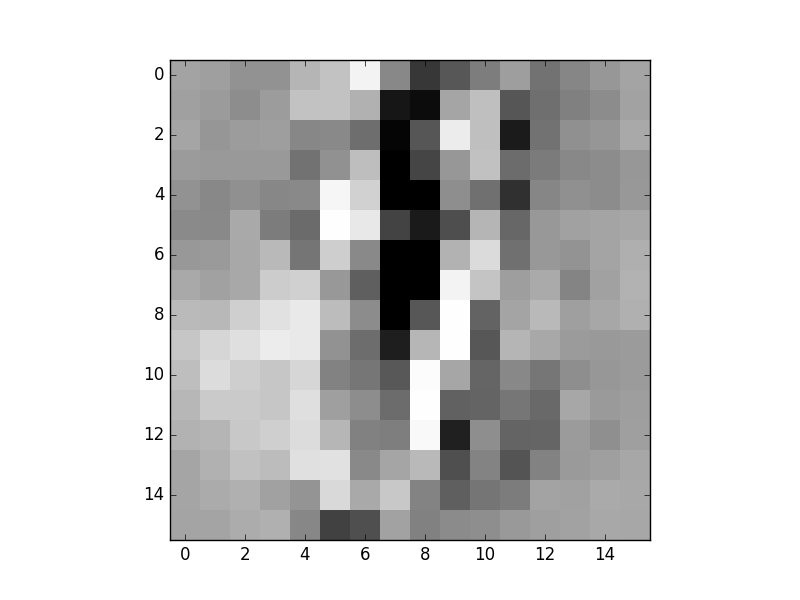}}&
\resizebox{0.3\columnwidth}{!}{\includegraphics[trim={4cm 0.5cm 4cm 0.5cm},clip]{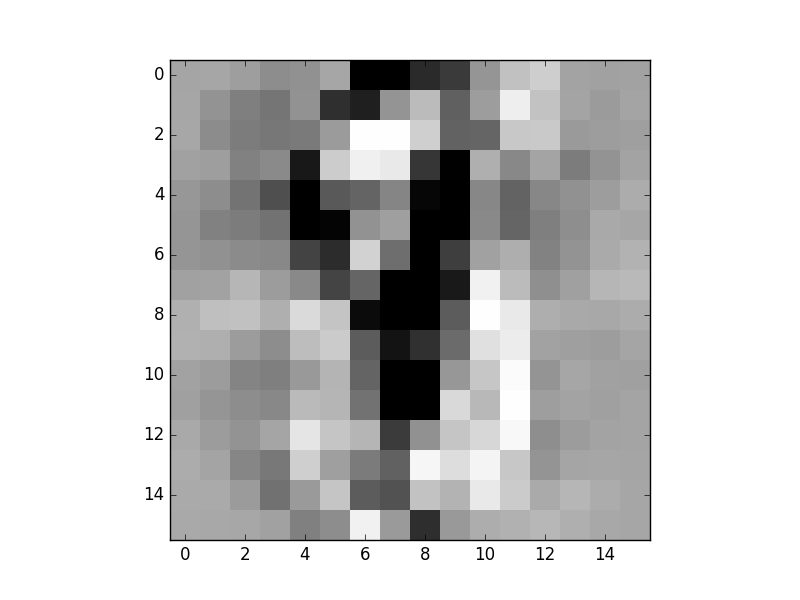}}\\
\resizebox{0.3\columnwidth}{!}{\includegraphics[trim={4cm 0.5cm 4cm 0.5cm},clip]{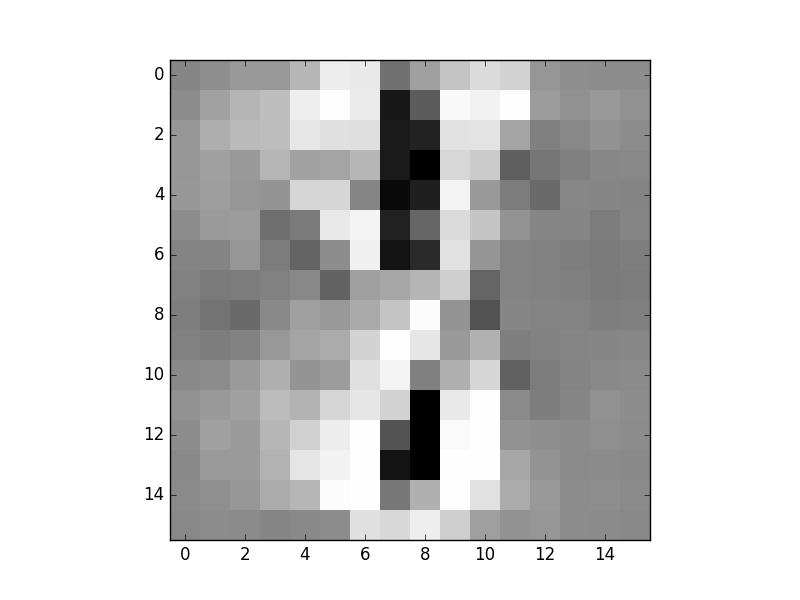}}&
\resizebox{0.3\columnwidth}{!}{\includegraphics[trim={4cm 0.5cm 4cm 0.5cm},clip]{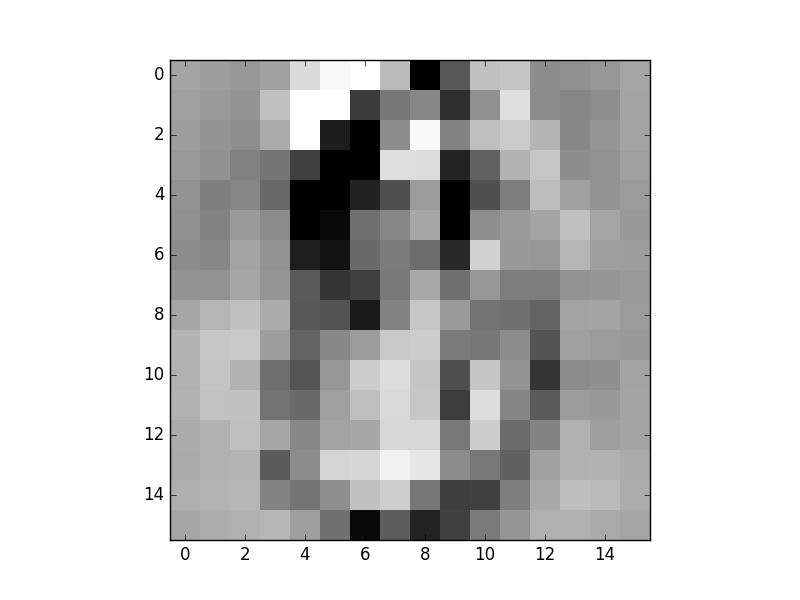}}&
\resizebox{0.3\columnwidth}{!}{\includegraphics[trim={4cm 0.5cm 4cm 0.5cm},clip]{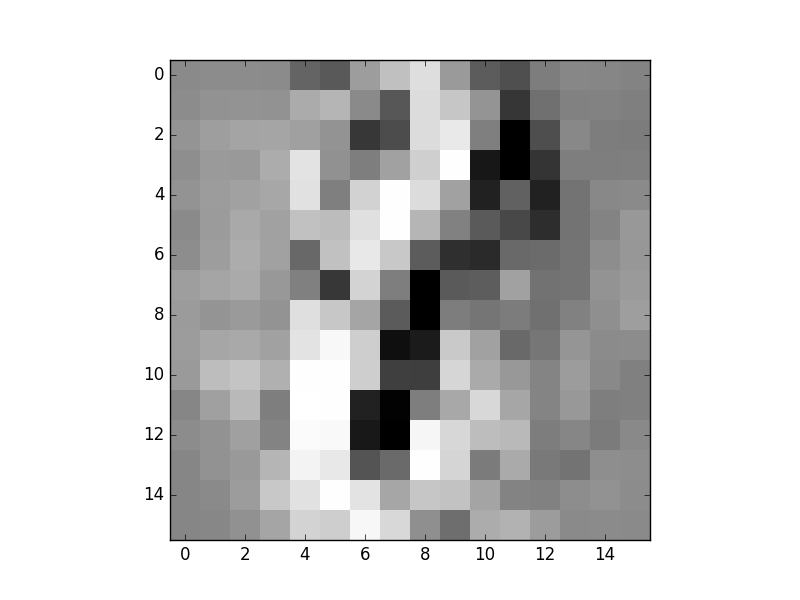}}\\
\end{tabular}
\\
\small Layer-by-Layer Unsupervised Autoencoder
&
\small Our Greedy Node-by-Bode Unsupervised Algorithm (GN)
\end{tabular}\vspace*{-10pt}
}
\caption{Features of nine nodes in the first internal layer for unsupervised pre-training.}
\label{fig:features-layers}
\end{center}
\vskip -0.2in
\end{figure*}
How one should train the internal layers? The two
popular approaches are:
(1) Each layer is 
an unsupervised nonlinear auto-encoder~\cite{CM88,bengio2007}; this approach
is appealing to build meaningful hierarchical
representations of the data in the internal layers. 
(2) Each layer is a supervised encoder; this approach primarly targets 
performance.
Deep learning enjoys success in several applications 
and hence considerable effort has been expended in optimizing the 
pre-training. The two main considerations are:

1. Pre-training time and training/test performance of the final solution.
The greedy layer-by-layer pre-training is significantly
more efficient than full backpropagation, and appears to be better
at avoiding bad local minima~\cite{erhan2010}. Our algorithms will show an order of
magnitude speed gain over greedy layer-by-layer pre-training.

\begin{figure*}[!ht]
\vskip 0.2in
\begin{center}
{\begin{tabular}{cc}
\setlength{\tabcolsep}{3pt}
\begin{tabular}{ccc}
\resizebox{0.3\columnwidth}{!}{\includegraphics[trim={4cm 0.5cm 4cm 0.5cm},clip]{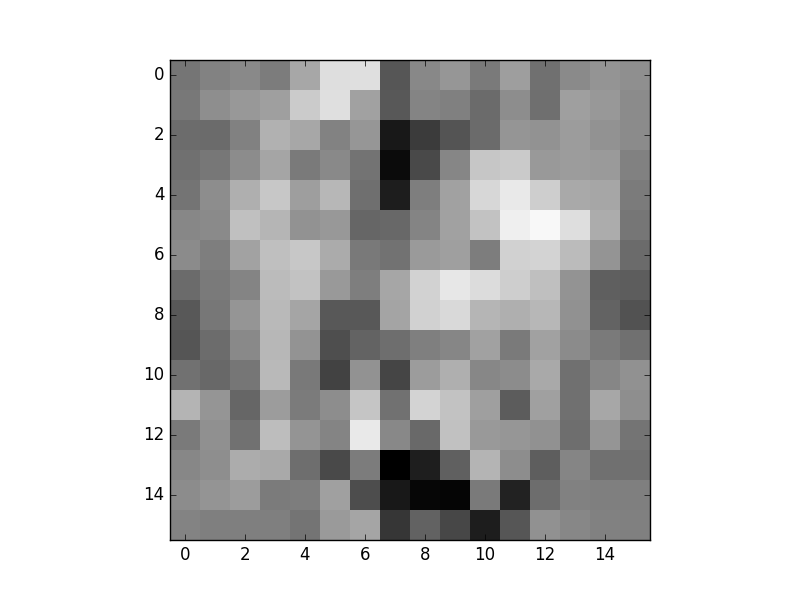}}&
\resizebox{0.3\columnwidth}{!}{\includegraphics[trim={4cm 0.5cm 4cm 0.5cm},clip]{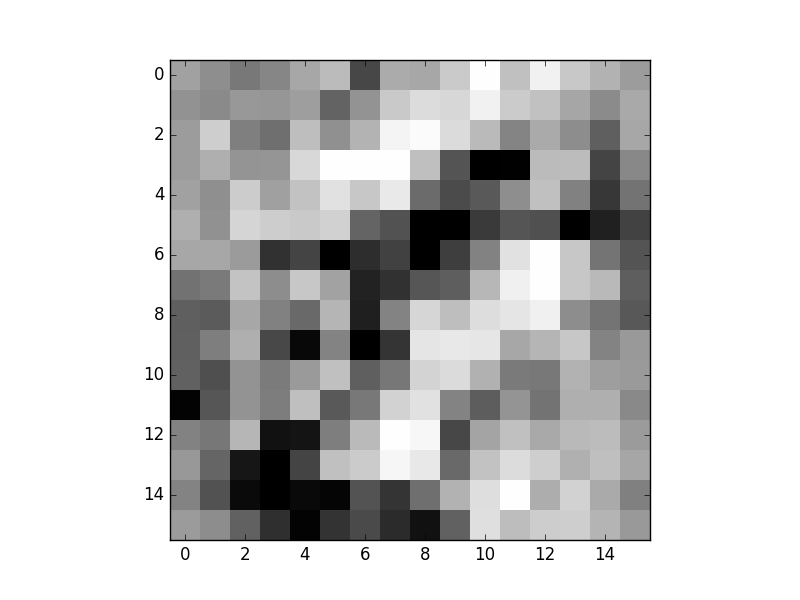}}&
\resizebox{0.3\columnwidth}{!}{\includegraphics[trim={4cm 0.5cm 4cm 0.5cm},clip]{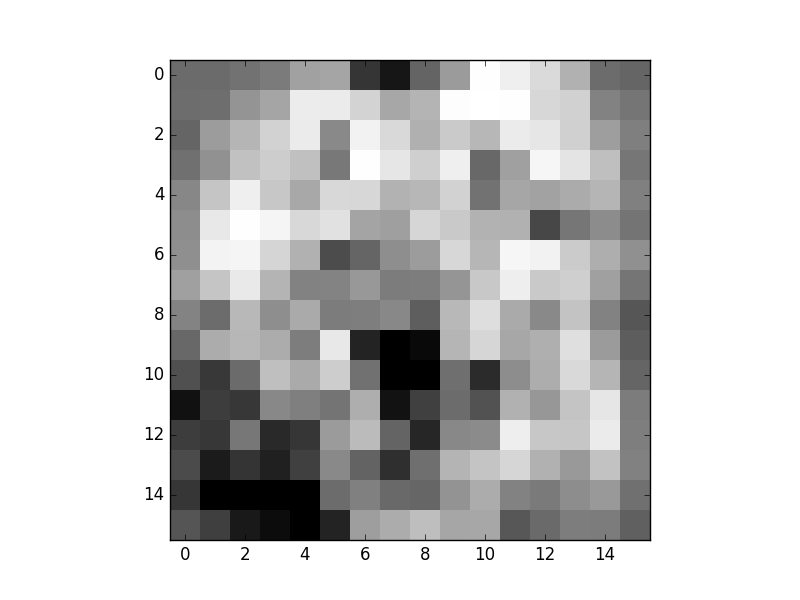}}\\
\resizebox{0.3\columnwidth}{!}{\includegraphics[trim={4cm 0.5cm 4cm 0.5cm},clip]{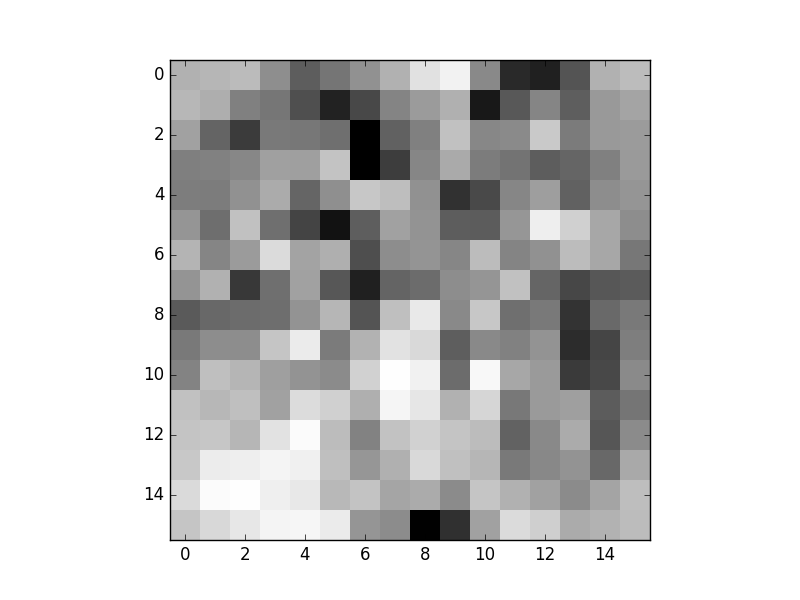}}&
\resizebox{0.3\columnwidth}{!}{\includegraphics[trim={4cm 0.5cm 4cm 0.5cm},clip]{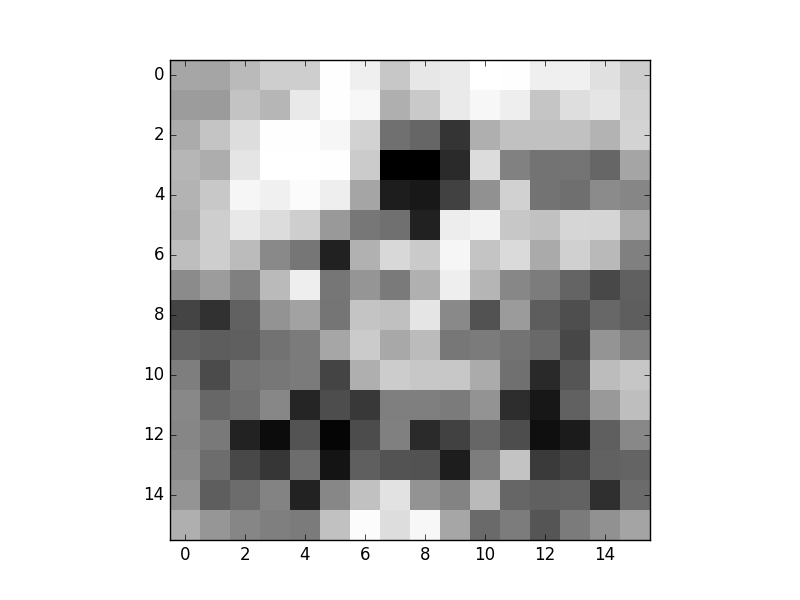}}&
\resizebox{0.3\columnwidth}{!}{\includegraphics[trim={4cm 0.5cm 4cm 0.5cm},clip]{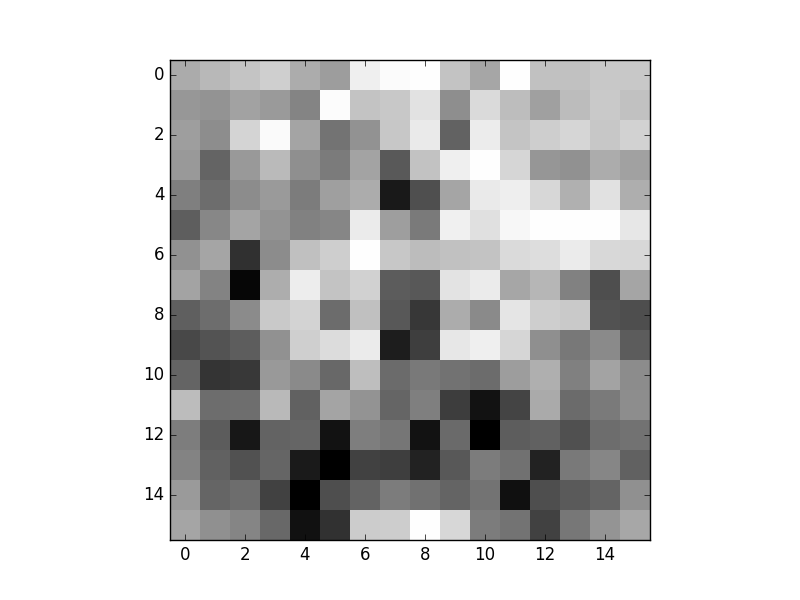}}\\
\resizebox{0.3\columnwidth}{!}{\includegraphics[trim={4cm 0.5cm 4cm 0.5cm},clip]{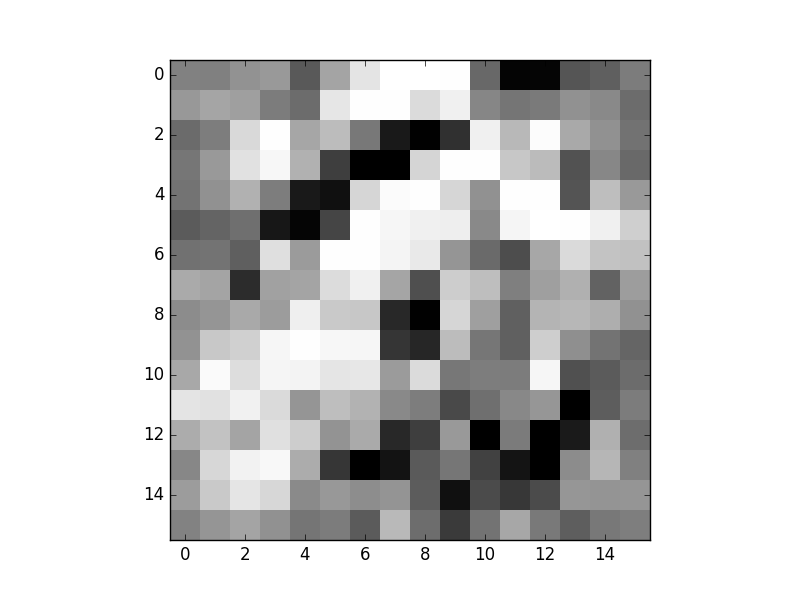}}&
\resizebox{0.3\columnwidth}{!}{\includegraphics[trim={4cm 0.5cm 4cm 0.5cm},clip]{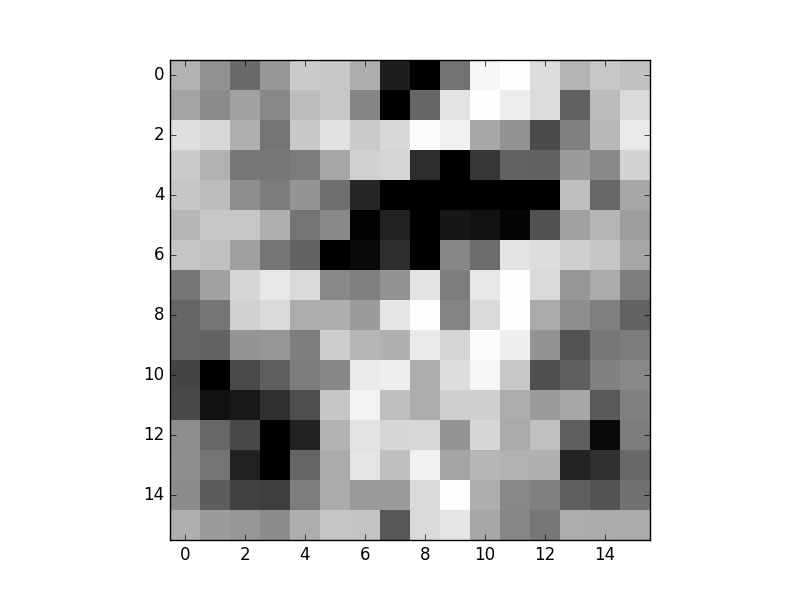}}&
\resizebox{0.3\columnwidth}{!}{\includegraphics[trim={4cm 0.5cm 4cm 0.5cm},clip]{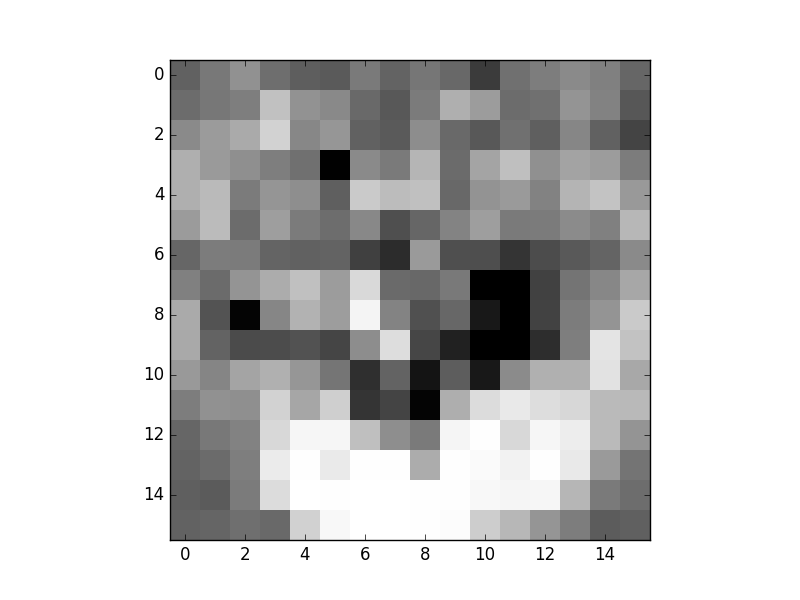}}\\
\end{tabular}
&
\setlength{\tabcolsep}{3pt}
\begin{tabular}{ccc}
\resizebox{0.3\columnwidth}{!}{\includegraphics[trim={4cm 0.5cm 4cm 0.5cm},clip] {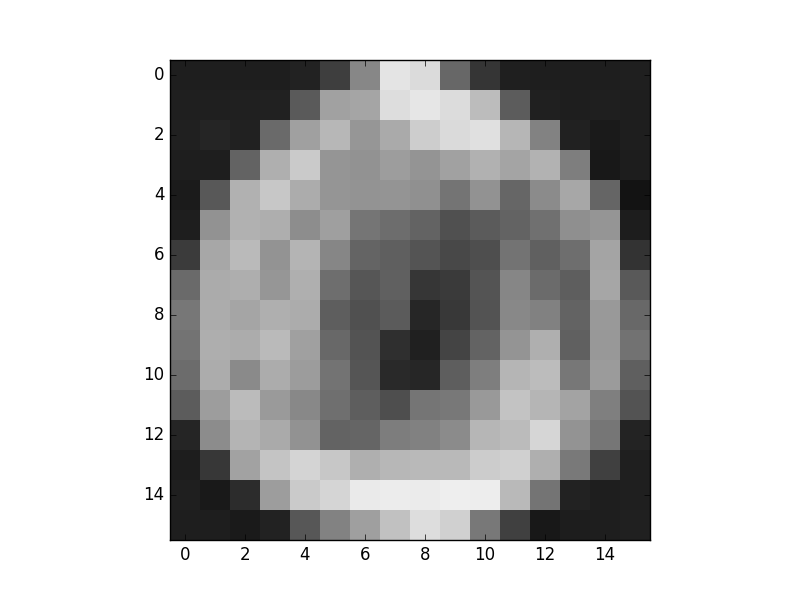}}&
\resizebox{0.3\columnwidth}{!}{\includegraphics[trim={4cm 0.5cm 4cm 0.5cm},clip]{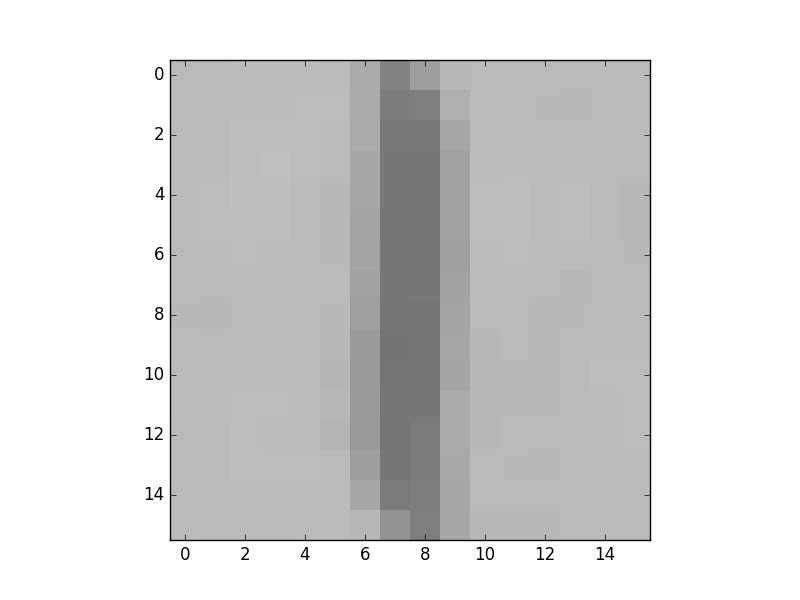}}&
\resizebox{0.3\columnwidth}{!}{\includegraphics[trim={4cm 0.5cm 4cm 0.5cm},clip]{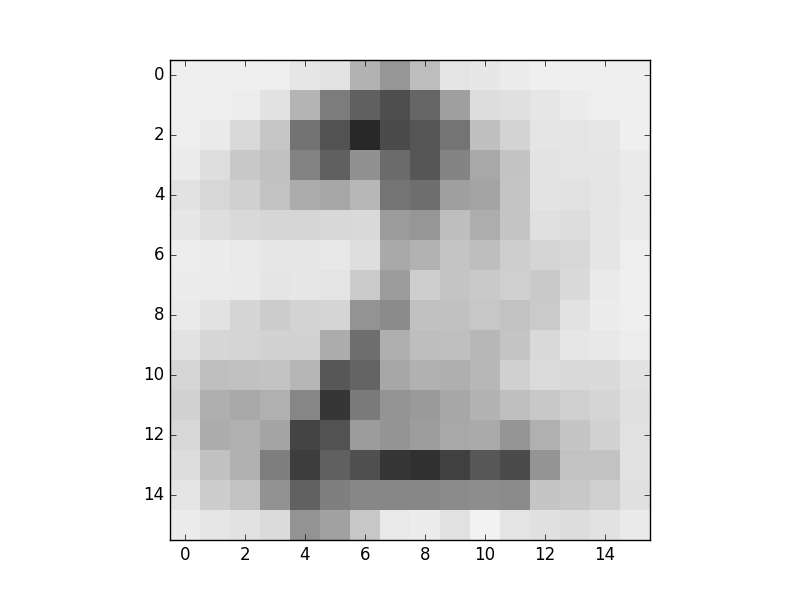}}\\
\resizebox{0.3\columnwidth}{!}{\includegraphics[trim={4cm 0.5cm 4cm 0.5cm},clip]{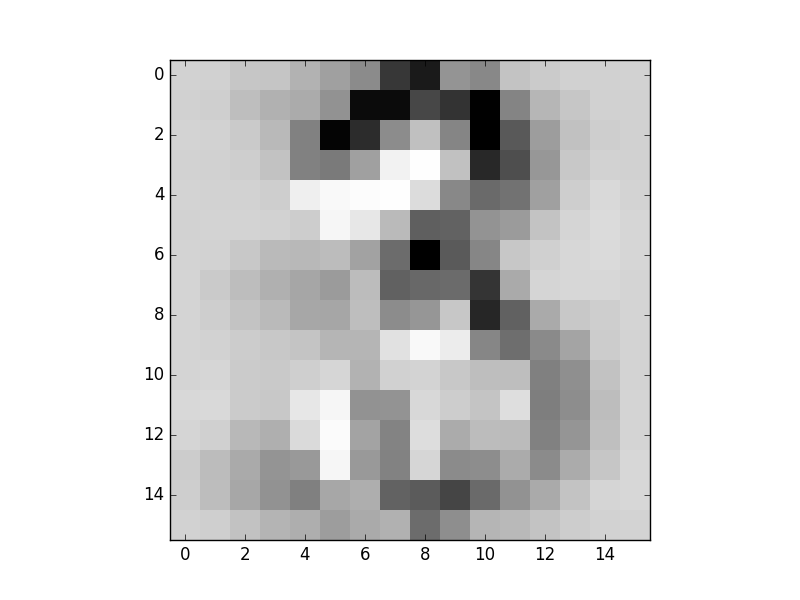}}&
\resizebox{0.3\columnwidth}{!}{\includegraphics[trim={4cm 0.5cm 4cm 0.5cm},clip]{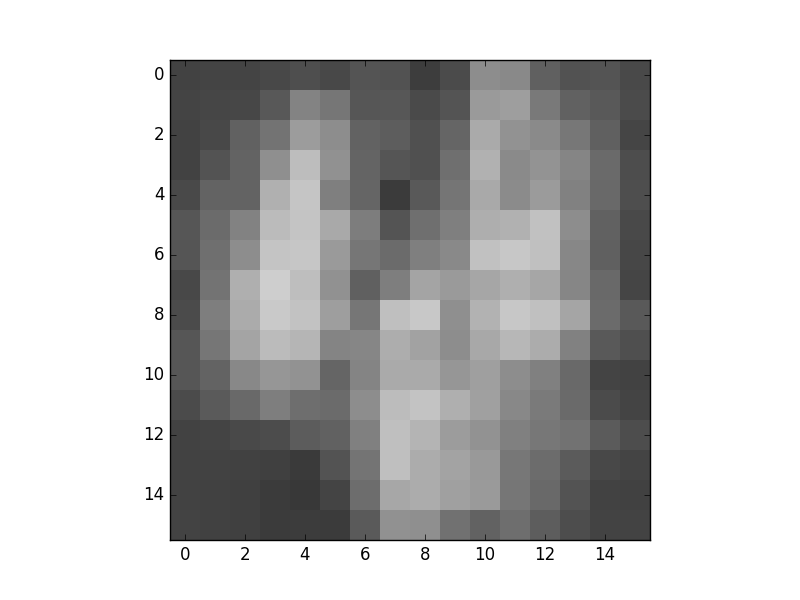}}&
\resizebox{0.3\columnwidth}{!}{\includegraphics[trim={4cm 0.5cm 4cm 0.5cm},clip]{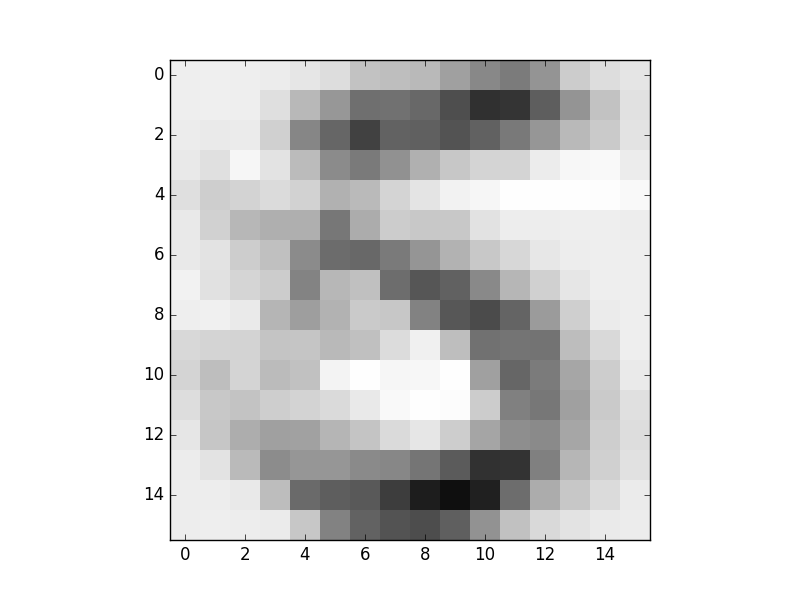}}\\
\resizebox{0.3\columnwidth}{!}{\includegraphics[trim={4cm 0.5cm 4cm 0.5cm},clip]{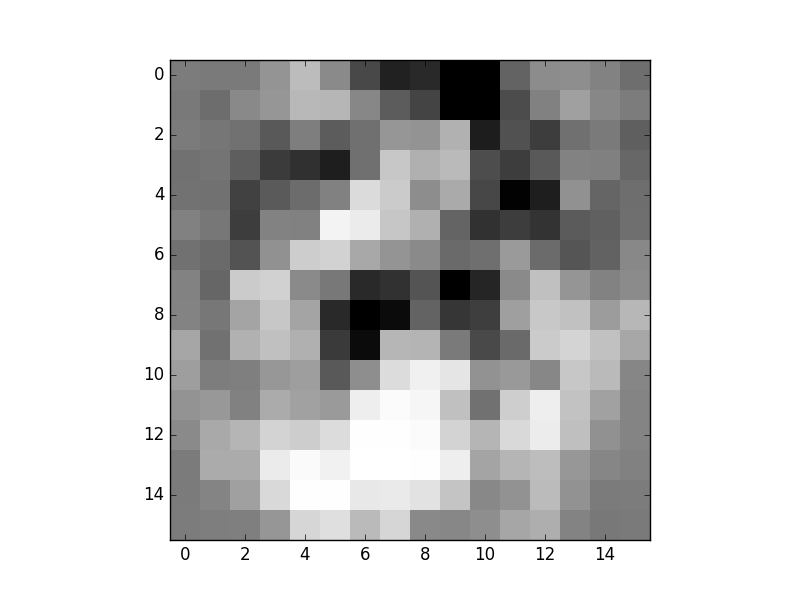}}&
\resizebox{0.3\columnwidth}{!}{\includegraphics[trim={4cm 0.5cm 4cm 0.5cm},clip]{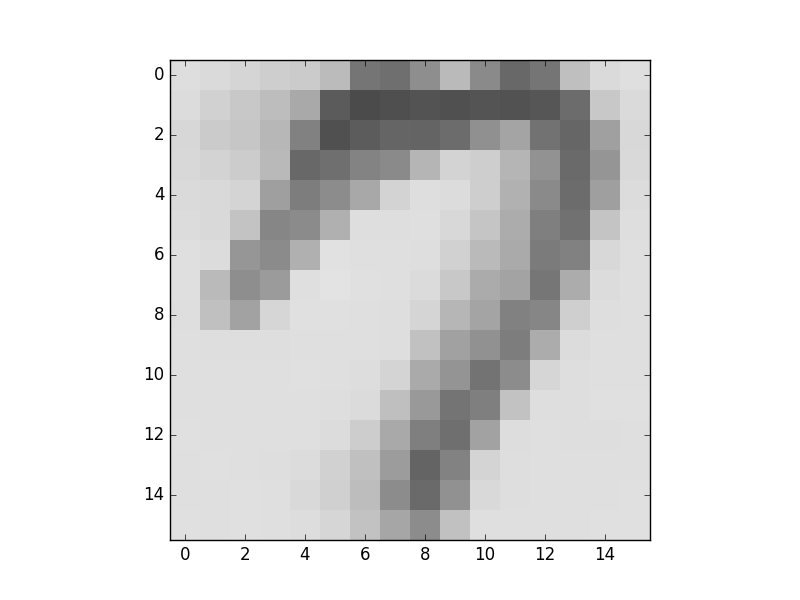}}&
\resizebox{0.3\columnwidth}{!}{\includegraphics[trim={4cm 0.5cm 4cm 0.5cm},clip]{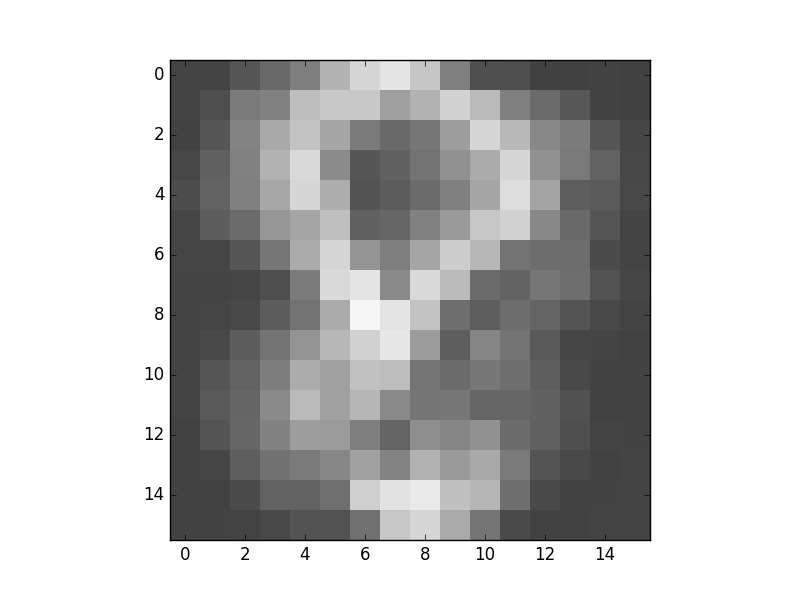}}\\
\end{tabular}
\\
\small Layer-by-Layer Supervised Encoder
&
\small Our Greedy Node-by-Node Supervised Algorithm (GCN)
\end{tabular}\vspace*{-10pt}
}
\caption{Features of nine nodes in the first internal layer for supervised pre-training.}
\label{fig:sup-features-layers}
\end{center}
\vskip -0.2in
\end{figure*}

2. Interpretability of the feature representations in the
internal layers. We use the USPS
digits data (10 classes)
as a strawman benchmark to illustrate our approach.
The  weights going into each node of the first layer
identify the pixels 
contributing to the ``high-level'' feature represented by that hidden node. 
Figure~\ref{fig:features-layers} compares our features with the 
layer-by-layer features in the unsupervised setting, and 
Figure~\ref{fig:sup-features-layers} compares the features in the 
supervised setting. The layer-by-layer features do not capture
the essence of the digits as do our features. This has to do with
the simultaneous training of all the hidden nodes in the layer. 
Our approach can be viewed as a nonlinear
extension of PCA, which "greedily" constructs each linear features.
(In the supplementary material, we show the features captured by linear PCA; 
they are comparable to our features.)

\begin{figure}[ht]
{\tabcolsep5pt
\begin{tabular}{ccccc}
\resizebox{0.15\columnwidth}{!}{\includegraphics*{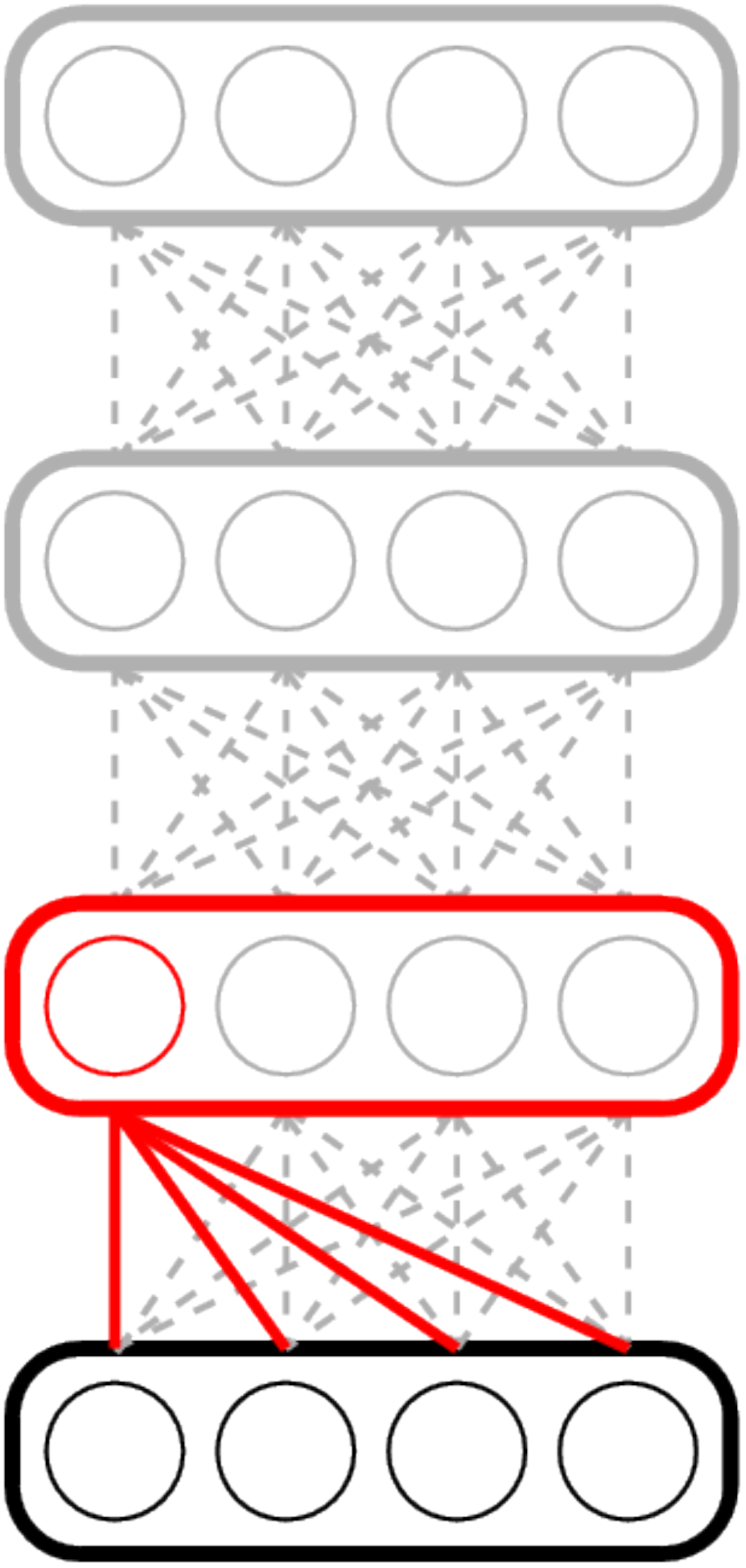}}&
\resizebox{0.15\columnwidth}{!}{\includegraphics*{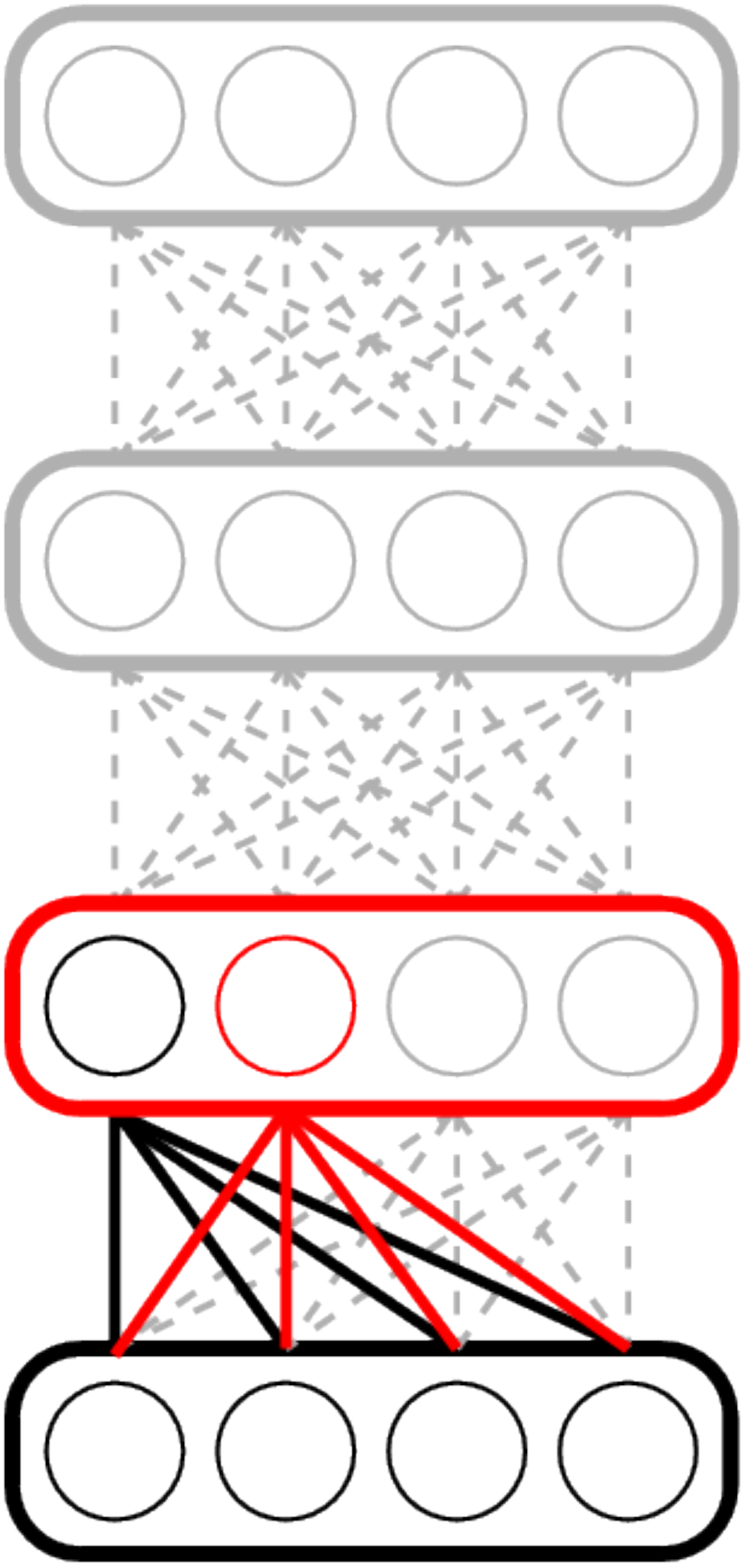}}&
\resizebox{0.15\columnwidth}{!}{\includegraphics*{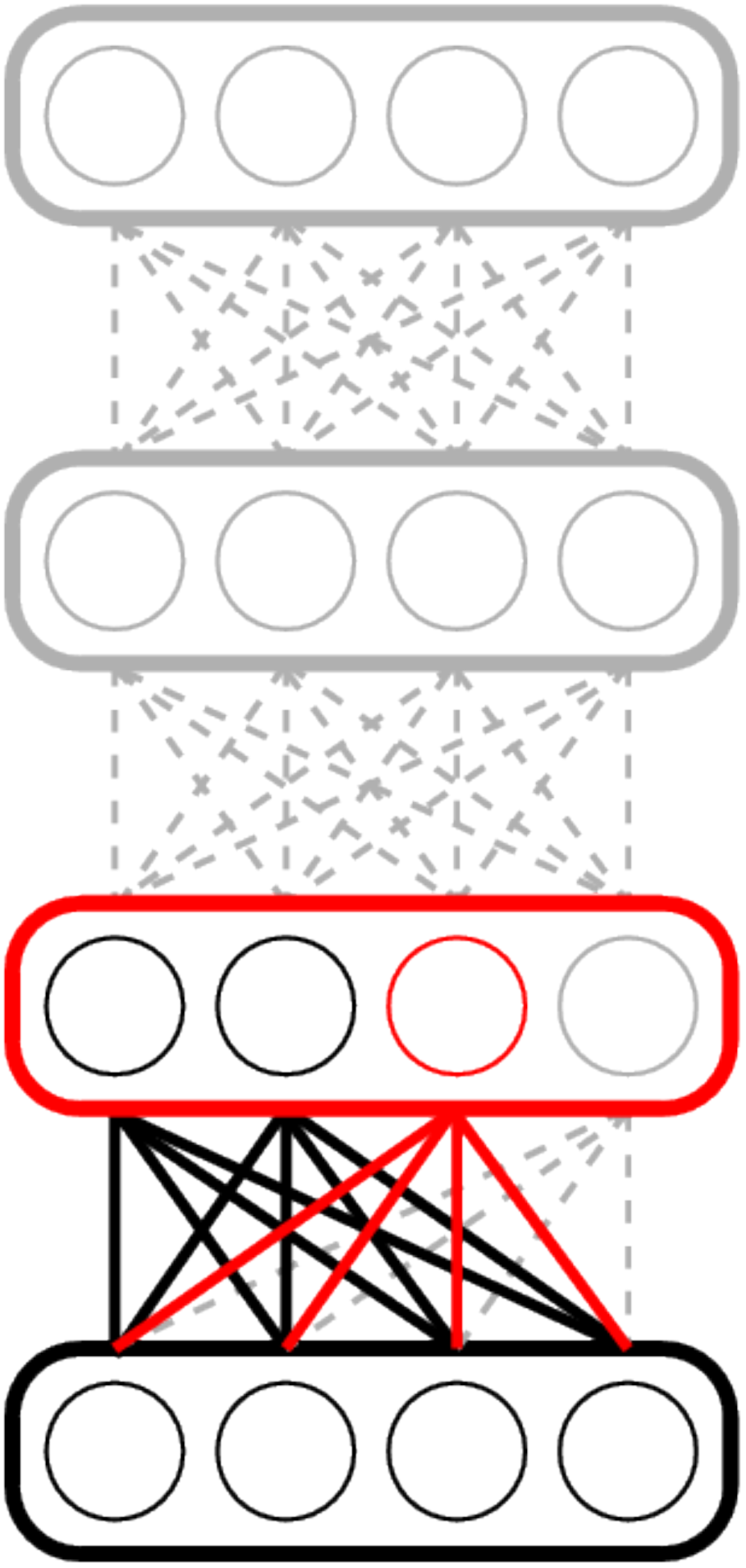}}&
\resizebox{0.15\columnwidth}{!}{\includegraphics*{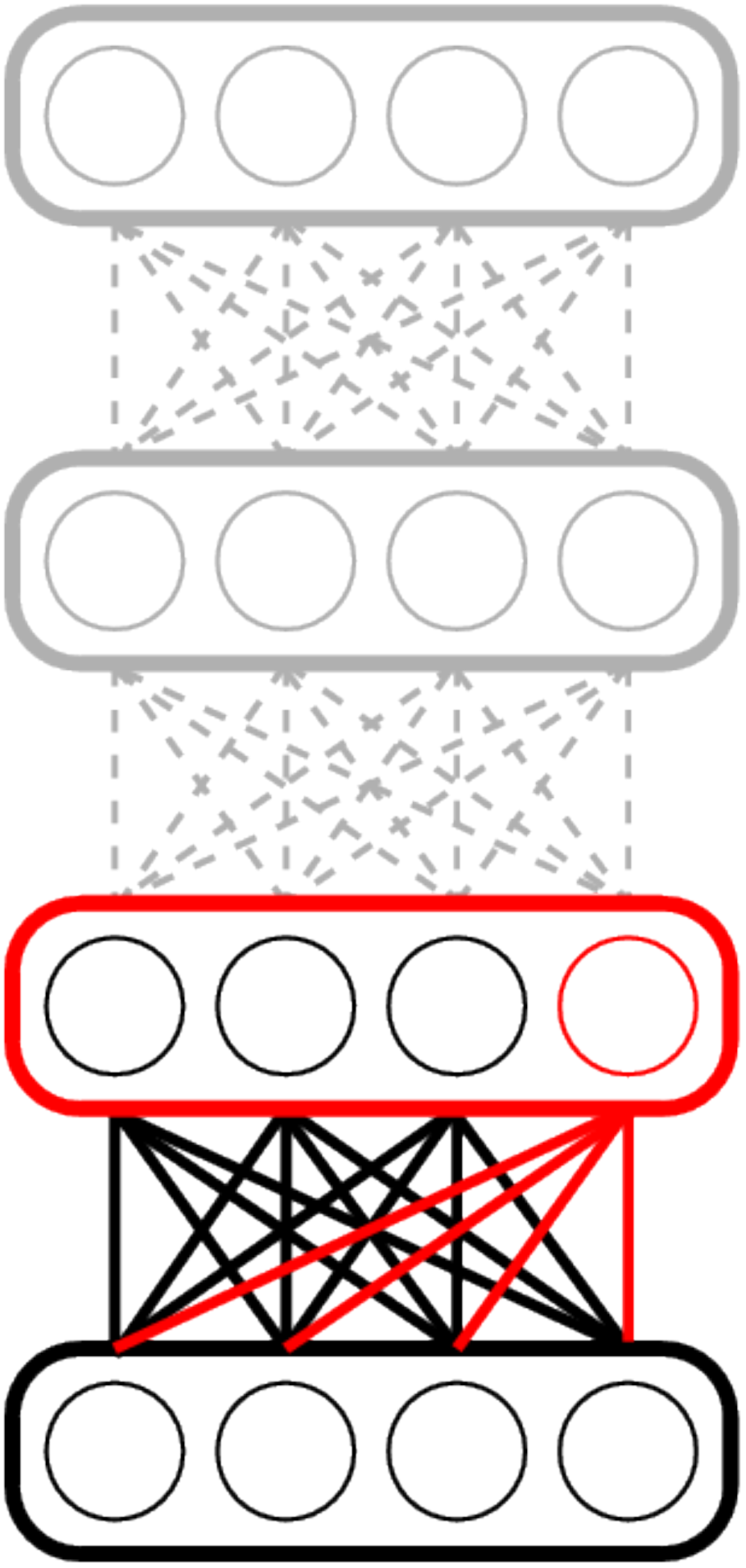}}&
\resizebox{0.15\columnwidth}{!}{\includegraphics*{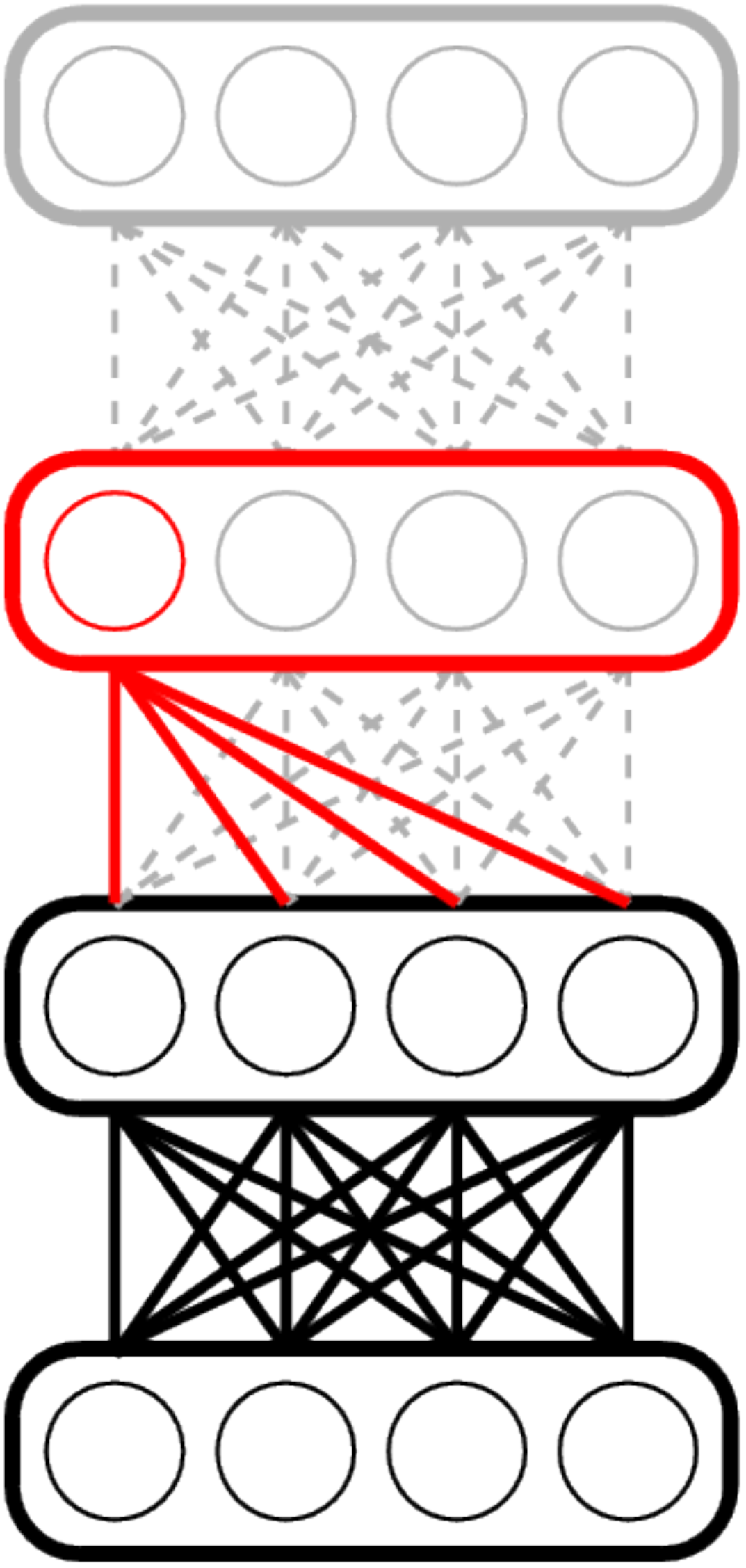}}
\\
\small \math{\matW^{(1)}_1}&
\small \math{\matW^{(1)}_2}&
\small \math{\matW^{(1)}_3}&
\small \math{\matW^{(1)}_4}&
\small \math{\matW^{(2)}_1}
\end{tabular}\vspace*{-5pt}
}
\caption{Node-by-node greedy  deep learning algorithm. In each step 
the weights feeding into one node are learned (red).
\label{fig:nodes}}
\end{figure}

\paragraph{Greedy Node-by-Node Pre-Training.}
The thrust of our approach is to learn the weights into each 
\emph{node} in a sequential greedy manner:  
\emph{greedy-by-node} (GN) for the unsupervised setting and
\emph{greedy-by-class-by-node} (GCN) for the supervised setting. 
Figure~\ref{fig:nodes} illustrates the first 5 steps for a network.
Our approach mimics a human who hardly 
builds all features
at once from all objects. 
Instead, features are built sequentially, while processing
the data.
Our algorithm learns one feature at a time, using a part of 
the data to learn each feature. Our contributions are 
\begin{enumerate}\itemsep-4pt
\vspace*{-10pt}
\item The specific algorithm to train each internal node.
\item How to select the training data for each internal node.
\vspace*{-20pt}
\end{enumerate}
We do \emph{not} improve the accuracy of deep 
learning. Rather, we improve efficiency and the  
interpretability of features, while \emph{maintaining} 
accuracy. A standard deep learning algorithm
uses every data point to process every weight in the network. Our
algorithm uses only a subset of the data to process a particular weight.
By training each node using "relevant" data, our algorithm 
produces more interpretable features. 
Our algorithm gets more
intuitive features, in 
the unsupervised and supervised setting
(Figures~\ref{fig:features-layers}~and~\ref{fig:sup-features-layers}).

\paragraph{Related Work.}
To help motivate our approach, it helps to 
start back at the very beginning of neural networks, with~\citet{rosenblatt1958}
 and \citet{widrow1960}. 
They introduced the adaline, the adaptive linear
(hard threshold element), and the combination of
multiple elements came in~\citet{hoff1962}, the madaline (the precursor
to the multilayer perceptron).
Things cooled off a little because fitting data with multiple 
hard threshold elements was a combinatorial nightmare. There is no doubt
softening the hard threshold to a sigmoid and the arrival of a new
efficient training algorithm, backpropagation~\cite{rumelhart1986}, was a 
huge part of the resurgence of neural networks in the 1980s/1990s. But,
again, neural networks receded, taking a back seat to
modern techniques like the
support vector machine~\cite{vapnik2000}. In part, this was due to the 
facts that multilayer feedforward networks were still hard to train iteratively
due to convergence issues and multiple
local minima~\cite{gori1992,fukumizu2000}, are
extremely powerful~\cite{hornik1989} and easy to overfit
to data. 
For these reasons, and 
despite the complexity theoretic advantages of deep networks
(see for example the short discussion in \cite{bengio2007}),
application of neural networks was limited
mostly to shallow two layer networks.
Multi-layer (deep) neural networks are back in the guise 
of deep learning/deep networks, and again because of 
a leap in the methods used to train the network \cite{hinton2006}.
In a nutshell, rather that address the full problem of learning the
weights in the network all at once, train each layer of the
network sequentially. In so doing, training becomes manageable
\cite{hinton2006,bengio2007}, the local 
minima problem when training a single layer is significantly diminished
as compared to the whole network and the restriction to layer by layer learning
reigns in the power of the network, 
helping with regularizing it\cite{erhan2010}. A side benefit has also emerged,
which is that each layer successively has the potential to learn
hierarchical representations~\cite{erhan2010,lee2009deep}.
As a result of these algorithmic advances, deep networks have found a
host of modern applications, ranging from 
sentiment classification~\cite{glorot2011},
to audio~\cite{lee2009}, to signal and information processing~\cite{yu2011deep}, 
to speech~\cite{deng2013}, and even to the 
unsupervised and transfer settings~\cite{bengio2012}.
Optimization of such deep networks is also an active area, 
for example~\cite{ngiam2011,martens2010}.

Most work has focused on better representations of the input data. Besides 
the original deep belief network~\cite{hinton2006} and 
autoencoder~\cite{bengio2007}, the stacked
denoising autoencoder~\cite{vincent2010stacked,VincentPLarochelleH2008} 
has been widely used as a variant to the classic autoencoder, where
corrupted data is used in pre-training.
Sparse encoding~\cite{boureau2008sparse,poultney2006efficient} has also
been used to prevent the system from activating the 
same subset of nodes constantly~\cite{poultney2006efficient}. 
This approach has been shown capable of learning local and meaningful features,
however the efficiency is worse.
For images, convolutional neural networks~\cite{lecun1995convolutional} 
and the deep convolutional neural network ~\cite{krizhevsky2012imagenet} 
are widely used, however the computation cost is significantly more, 
and the feature learning is based on the local filter defined by the modeler. 
Our methods apply to a general deep network.
Our algorithms can be roughly seen as simultaneously 
clustering the data while extracting the features. 
Deep networks have been used for unsupervised clustering \cite{chen2015deep}
and clustering has been used in classic deep networks
by~\citet{weston2012deep} which are targeted for semi-supervised learning. 
Using clusters from  \math{K}-means
to train a deep network was reported~\cite{faraoun2006neural},
which improves the training speed while ignoring some data - not 
recommended in the supervised setting with scarce data.
Pre-clustering may have efficiency-advantages in large-scale systems, but
it may not be an effective way 
to learn good representations~\cite{coates2012learning}. 

In this paper, we are proposing a 
new algorithmic enhancement to the deep network which is 
to consider each node separately. It not only explicitly achieves the 
sparsity of node activation but also requires 
much shorter computation time. We have found no such approaches in the
literature.  

%% file: method.tex
\section{Greedy Node-by-Node Deep Learning}
The basic step for pre-training is to train a two layer network. 
The network is trained to reproduce the input (unsupervised) or the final 
target (supervised). The two algorithms are very similar in structure. 
For concreteness,
we describe unsupervised pre-training (auto-encoding).

In a classic auto-encoder with one hidden layer using 
stochastic gradient descent (SGD), 
the number of operations (fundamental arithmetic operations) 
$p$ for one training example
is:
\mldc{
\arraycolsep2pt
\begin{array}{rclcl}
p &=&(2d_1d_2 + d_1+2d_2)&+&(3d_1d_2 + 2d_1+3d_2)\\
&&\text{\small(Forward Propagation)}&&\text{\small(Backpropagation)} \\ 
&=& 5d_1d_2 + 3d_1 + 5d_2,
\end{array}
\label{AE_eqn_std}
}
where $d_1$ is the dimension of the input layer,
($d_1+1$ including the bias) and 
$d_2$ is the dimension of the second layer. 
Forward propagation  
computes the output and backpropagation computes the
gradient of the loss w.r.t. the weights (see~\citet[Chapter 7]{malik173}).
We use the Euclidean distance between the 
reconstructed input \math{\hat\xx} and the original input \math{\xx} 
(auto-encoder target) as loss,
\mandc{\text{loss}=\norm{\xx-\hat\xx}^2.}   
For \math{N} training example and \math{E} epochs of 
SGD,  the total running time is $O(NpE)=O(NEd_1d_2)$.  

\begin{figure}[t]
\vskip 0.2in
\begin{center}
\scalebox{0.75}{
\begin{tikzpicture}[x=0.9cm,y=1.25cm]
\foreach\x[count=\i] in {1,...,5}
{
\node[circle,draw,inner sep=5pt](F\i) at (\i,0){};
}
\foreach\x[count=\i] in {1,...,4}
{
\node[circle,draw,inner sep=5pt](S\i) at (\i,1.5){};
}
\foreach\x[count=\i] in {1,...,6}
{
\node[circle,draw,inner sep=5pt](T\i) at (\i,3){};
}

\draw[rounded corners=7pt,line width=2pt]
($(F1.225)+(-0.2,-0.2)$)rectangle($(F5.45)+(0.2,0.2)$);
\draw[rounded corners=7pt,line width=2pt]
($(S1.225)+(-0.2,-0.2)$)rectangle($(S4.45)+(0.2,0.2)$);
\draw[rounded corners=7pt,line width=2pt]
($(T1.225)+(-0.2,-0.2)$)rectangle($(T6.45)+(0.2,0.2)$);

\foreach\x in{1,...,5}
\foreach\y in{1,...,4}
\draw[line width=1.5pt,dashed,lightgray](F\x)--(S\y);

\foreach\x in{1,...,4}
\foreach\y in{1,...,6}
\draw[line width=1.5pt,dashed,lightgray](S\x)--(T\y);

\foreach\x in{1,...,5}
\draw[line width=1.25pt](F\x)--(S1);

\foreach\x in{1,...,6}
\draw[line width=1.25pt](S1)--(T\x);

\foreach\x in{1,...,5}
\draw[red,line width=1.75pt](F\x)--(S2);

\foreach\x in{1,...,6}
\draw[red,line width=1.75pt](S2)--(T\x);

\node[circle,draw,inner sep=5pt,fill] at (S1){};
\node[circle,draw=red,inner sep=5pt,pattern=north east lines,pattern color=red,line width=1.5pt] at (S2){};

\end{tikzpicture}
}
\caption{Greedy training of a 2-layer network}
\label{greedy_network}
\end{center}
\vskip -0.2in
\end{figure}
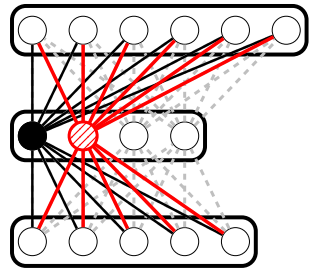

In our algorithm, the basic step is also to train a 
2-layer network. However, we train each node sequentially in 
a greedy fashion as illustrated in Figure~\ref{greedy_network}. 
The red (middle) layer is being trained (it has dimension \math{d_2}). 
The inputs come from the outputs of the previous layer, 
having dimension
\math{d_1}. The output-dimension is also \math{d_1} 
(auto-encoder). We use linear output-nodes and SGD optimizing the auto-encoder
(the algorithm is easy to adapt to sigmoid output-nodes).

The standard layer-by-layer algorithm trains all the weights at the same time, 
using the whole data set. 
We are going to use a fraction of the data to learn one feature at a time; 
different features are learned on different data.
So, the training of each layer is done in multiple stages. 
At each stage, we only update the weights corresponding to one node.
After all the features are obtained (all the weights learned) for a layer,
a forward propagation with all data computes the outputs of the layer,
for use in training the next layer (as in standard pre-training).
To make this greedy learning algorithm work, we must address three questions:
\begin{enumerate}\itemsep-4pt
\vspace*{-8pt}
\item How to learn features sequentially? 
\item How to distribute the training data into each node? 
\item How to obtain non-overlapping features?
\vspace*{-8pt}
\end{enumerate}
Let us address the question 1, assuming we have already
created \math{d_2} subsets of the data of size \math{K}, 
\math{S_1,\ldots,S_{d_2}}. These subsets
need not be disjoint, but our discussion is for the case of disjoint subsets,
so \math{Kd_2=N}. If each node is fed a random subset of \math{K}
data points, then each node will be roughly learning the same feature,
that is close to the top principle component. We will address this
issue is questions 2 and 3 which encourage learning different features
by implementing a form of orthogonality and using different data for each node.

A simple idea which works but is inefficient: use data \math{S_i}
to train the weights into hidden node \math{i} (assuming weights into
nodes \math{1,\ldots,(i-1)} have been trained). The situation is illustrated
in Figure~\ref{greedy_network}.
The weights into and out of node 1 are fixed (black). 
Node 2 is being trained (the red weights into and out of node 2).
We use data \math{S_2} to train these red weights, so we forward propagate the
data through both nodes, but we only need to backpropagate through the 
red nodes. Effectively, we are forward propagating through a network of 
\math{i} hidden nodes, but we have \math{K} data points, so the run-time is
of just the forward propagations is
\mandc{
\sum_{i=1}^{d_2}O(KEd_1i)=O(KEd_1d_2^2).
}
Since \math{d_2} could be large, it is inefficient to have a 
quadratic in \math{d_2}
run-time. There is a simple fix which relies on the
fact that the outputs of the hidden layer are linearly summed
before feeding into the output layer. This means that after the weights for
node 1 are learned (and fixed), one can do a single forward 
propagation of all the data once through this node and store a running sum
of the signal being fed into each output-layer node, which is the sunning
contribution from all previously trained nodes in the hidden layer. 
This restores the linear dependence on \math{d_2} as 
in equation~\r{AE_eqn_std}.

\remove{
To solve the issue mentioned above, one solution seems to be using the first fraction to train the first node and then using the second fraction on the first two nodes (for forward propagation), but only updating corresponding weights for the second node. In this manner, the new feature learned relies on the previous knowledge, and the system can learn the generative representations similar to the standard pre-training algorithms. However, since for stage $i$ the forward propagation is applied on all the current $i$ ($ \leq d_2$) nodes, the total number of numerical operations in forward propagations turns out to be $\sum_{i=1}^{d_2} i = O(d_2^2) $, which makes the computation much slower, since the standard algorithms only have $O(d_2)$ operations (see equation~\ref{AE_eqn_std}). 
Our method utilizes the linearity of the transformation from the inner layer to the output layer (before the optional nonlinear activation function). Suppose $\boldsymbol{W}$ is the weight matrix between the inner layer to the output layer with dimension $d_1d_2$ (ignoring the bias node for the inner layer), and the outcome of the $d_2$ inner nodes is a vector $v$ of length $d_2$, the output after forward propagation is then $\boldsymbol{W}v$. From basic linear algebra one can find that $output=\boldsymbol{W}v = \sum_{j=1}^{d_2} W_{\cdot j}v_j$, where $W_{\cdot j}$ are the weights coming out of the inner node $j$ to the output layer and $v_j$ is the outcome value of this node ($v_j$ is a scalar value). This equation shows that the contributions from the inner nodes to the output layer are independent and additive. Assuming the output layer uses a linear activation function (identity function), the output of the network can be seen as the summation of all the $d_1$ dimension vectors generated from each inner node and in back propagation they are constrained by the summation, since the loss function is the Euclidean distance between the output and the target.  Consequently, it is not necessary to forward-propagate through all the first $i$ nodes at stage $i$, as the solution mentioned early. One can store the output layer from stage $i-1$ and then add it to the new output only through the node $i$ to get the same result. This process is equivalent to propagate through $i$ nodes but only update the weights for the newest one. In this way, for each stage, we only need to propagate through one inner node, which results in an overall number of operations of $O(d_2)$.
}

\emph{Distributing data into nodes:}
We now address the 2nd question. 
We propose two methods corresponding to the two new algorithms GN and GCN. 
There are many ways to extend these two basic methods, so we present only
the rudimentary forms. Each node \math{i}
is trained on subset \math{S_i}.
In the unsupervised setting (GN), we train node 1 to learn a
``global feature'', and use this single global feature to reconstruct all
the data. The reconstruction error from this single feature is then
used to rank the data. A small reconstruction error means the data point
is captured well by the current feature. Large reconstruction errors
mean a new feature is needed for those data. So the reconstruction error
can be used as an approximate proxy for partitioning the data into different
features: data with drastically
different reconstruction error correspond to different features,
so that data will be used to train its own feature.
After sorting the data according to reconstruction error, the first 
\math{N/(d_2-1)} will be used to train node \math{2}, the next 
\math{N/(d_2-1)} to train  node \math{3} and so on up to node \math{d_2}.
One may do more sophisticated things, like cluster the 
reconstruction errors into \math{d_2-1} disjoint clusters and use those 
clusters as the subsets \math{S_2,\ldots,S_{d_2}}. 
We present the simplest approach.

In the supervised setting (GCN), the only change is to modify the 
distribution of the data into nodes using the class labels. If there are
\math{c} classes, then \math{d_2/c} of the nodes will be dedicated to each 
class, and the data points in each class are distributed evenly among the
nodes dedicated to that class.

\emph{Ensuring Non-Overlapping Features.}
If every node is trained independently on all or a random
sample of the data, then each node will learn more-or-less the same feature.
Our method of distributing the data among the nodes to some extent 
breaks this tie. We also introduce an explicit coordination mechanism 
between the nodes which we call the \emph{amnesia} factor -- when training
the next node, how much of the prior training is ``forgotten''.
Recall that from the 
previous $i-1$ (already trained) nodes, we store 
the running contribution to the output.
SGD is applied based on the distance between the input 
and the sum of the current output and running stored value from the
previous \math{i-1} nodes. 
The running stored value can be viewed as a constraint on 
the training of node $i$, forcing the newly learned feature 
to be "orthogonal" to the previous ones. This is because the
previous features are encoded in the running stored value. The current value
produced by node \math{i} will try to contribute additional information
toward reconstruct the data in \math{S_i}.
Since, in our algorithms, the weights from the previous nodes
have been learned and fixed, due to their optimization, it may
prematurely saturate the output layer and make the new
\math{i}th node redundant. The amnesia factor gives the ability
to add just the right amount of coordination between
the training of node \math{i} and the nodes that have already been trained,
leading to stability of the features while at the same time
maintaining the ability to get non-redundant features.
Our implementation of amnesia is simple. The output value, \math{O_B},  
used for 
backpropagation to train the weights into node \math{i} are the
 stored (already learned) running output, \math{O_{(1:i-1)}}, 
\emph{scaled by the amnesia factor}
plus the output from the currently being trained node \math{O_i},
\mandc{O_B=A\cdot O_{(1:i-1)} + O_i.}
The amnesia factor controls how ``orthogonal'' each successive
feature will be to the previously trained features. 
A higher amnesia factor results in strongly coupled features
that are more orthogonal.
A zero amnesia factor means independent training of the nodes which is
likely to result in redundant features.
Here is a high-level summary of the full algorithm. Detailed pseudo-code is
presented in the supplementary materials.
\begin{algorithmic}[1]
\STATE Distribute the data into subsets $S_1,\ldots,S_{d_2}$
\STATE Train hidden node $i$ on data \math{S_i} using amnesia factor 
\math{A} 
\STATE Perform one forward propagation with \math{S_i} on nodes
\math{1,\ldots,i} after training, and add the output values to the running
output value.
\end{algorithmic}
\begin{theorem}
GN and GCN run in \math{O(NEd_1+Nd_1d_2)} time.
\end{theorem}
\remove{
\begin{theorem}
Let $d_1$, $d_2$ be the dimensions of input layer and inner layer (with no bias node in the inner layer), $N$ be the size of the data set, $c$ be the number of the classes in the data, $E$ be the number of epochs for the training, $F$ be the times of loss computation that needed to check the convergence, the running time of auto-encoder ($T_{AE}$), GN($T_{GN}$), GCN($T_{GCN}$) can be computed as follows:
\begin{itemize}
	\item $T_{AE} = O((NE+NF)d_1d_2)$ 
	\item $T_{GN} = O((NE+NF)d_1 + Nd_1d_2) $
	\item $T_{GCN} = O((NE+NF)d_1 + Nd_1d_2))$
\end{itemize}
, and unless $d_2 \gg E$, $T_{GCN} \approx 0.5T_{GN}$, otherwise $T_{GCN} = T_{GN} < T_{AE}$.
\end{theorem}   	
}
(The detailed derivation of the running time
is in the supplementary materials.)
The run-time of the classic deep-network algorithm
is \math{O(NEd_1d_2)}.  As \math{d_2} and \math{E} (the number of
iterations of SGD on each data point) increase, the efficiency gain increases,
and can be orders of magnitude.

%% file: expts.tex
\section{Results and Discussion}

We use a variety of data sets to compare our new algorithms
GN (unsupervised) and GCN (supervised)  against some standard deep-network
training algorithms. Our aim is to demonstrate that 
\begin{itemize}\itemsep-4pt
\vspace*{-8pt}
\item Features from GN and GCN are more interpretable.
\item The classification performance of our algorithms 
is comparable to layer-by-layer training, despite being 
orders of magnitude more efficient.
\vspace*{-8pt}
\end{itemize}

\begin{figure*}[ht]
\begin{center}
\begin{tabular}{c}
\begin{tabular}{ccc}
\setlength{\tabcolsep}{1.5pt}
\resizebox{0.575\columnwidth}{!}{\includegraphics[trim={2cm 0cm 3cm 0cm},clip]{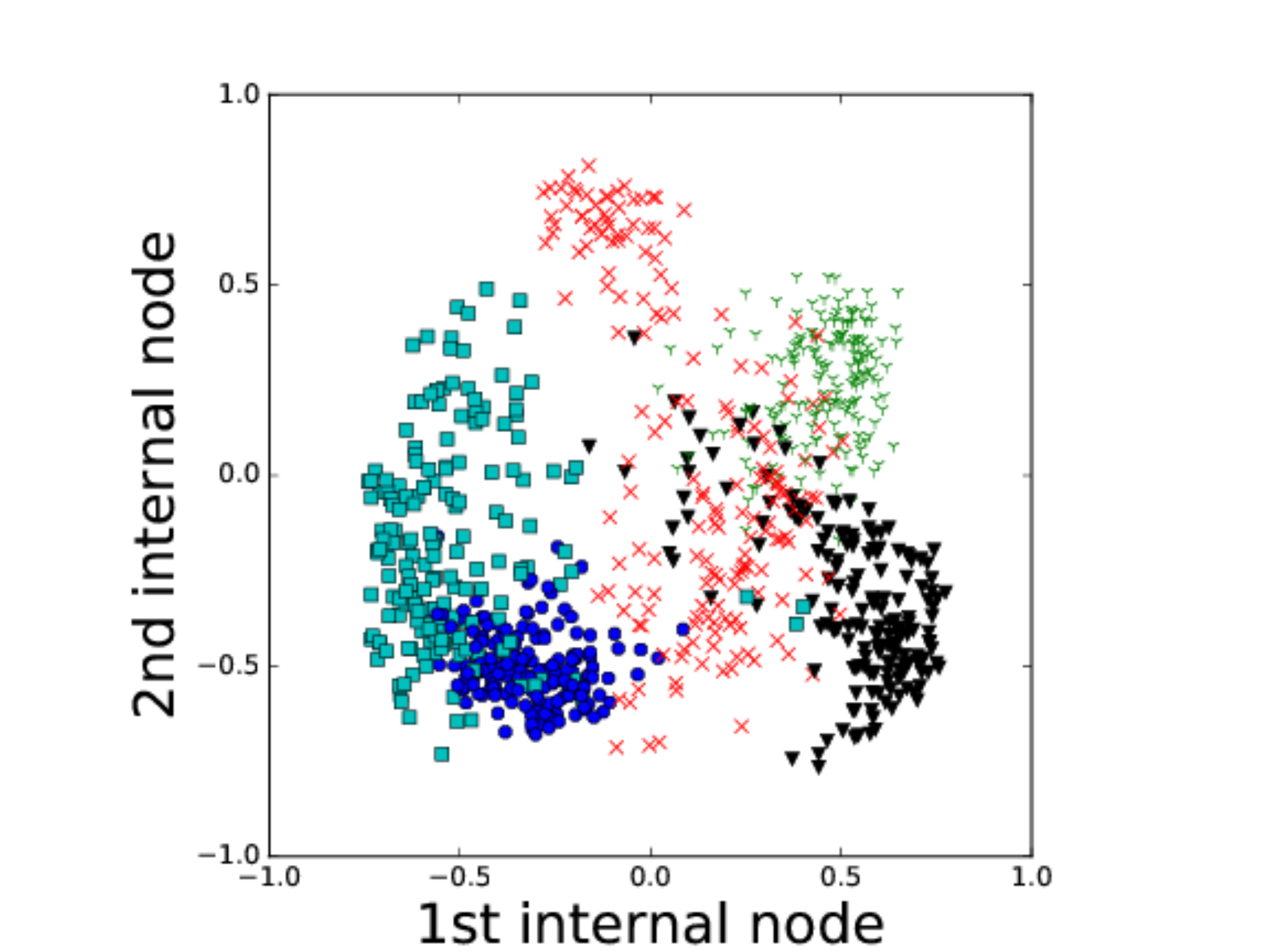}} &
\resizebox{0.575\columnwidth}{!}{\includegraphics[trim={2cm 0cm 3cm 0cm},clip]{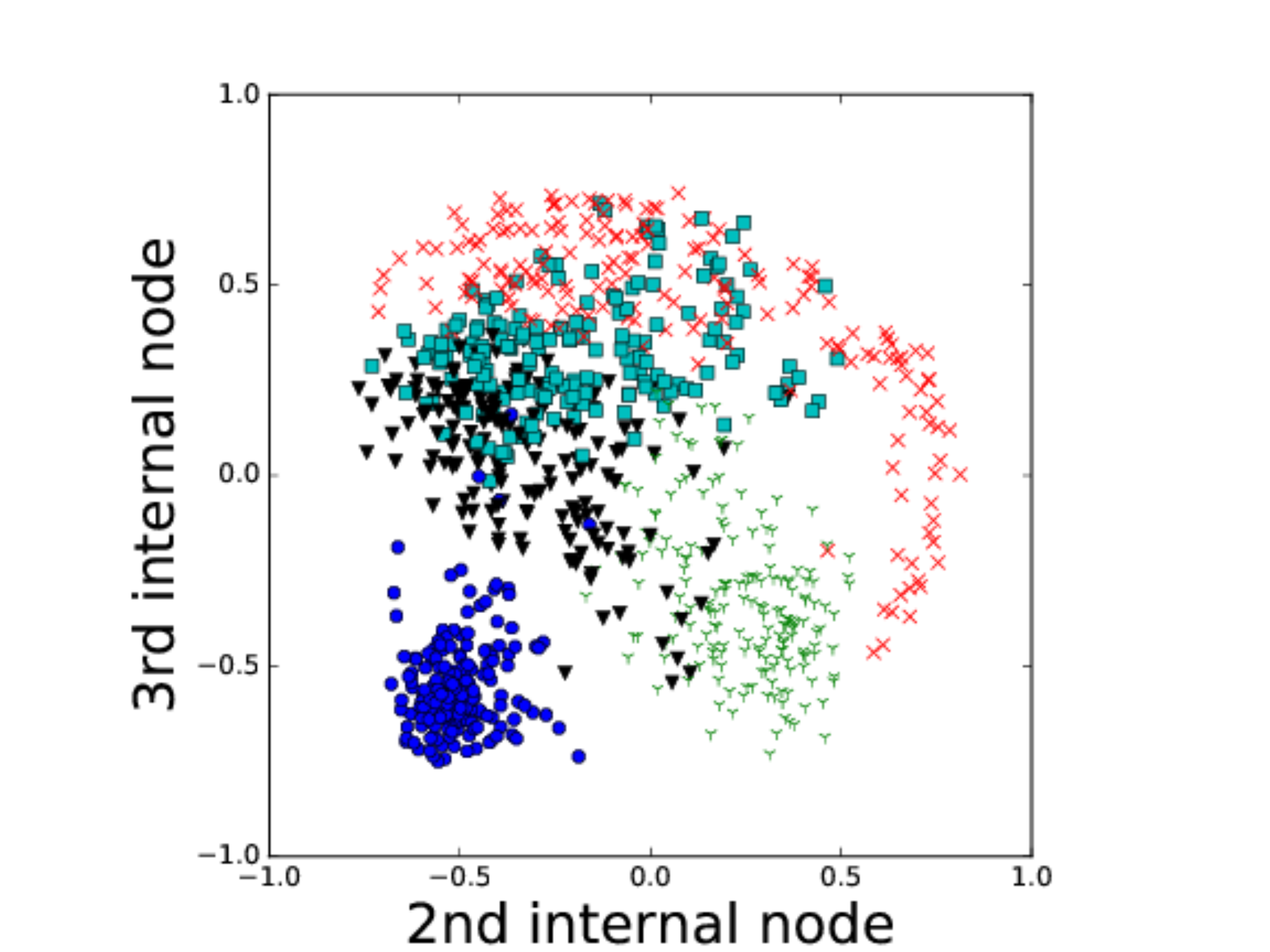}} &
\resizebox{0.575\columnwidth}{!}{\includegraphics[trim={2cm 0cm 3cm 0cm},clip]{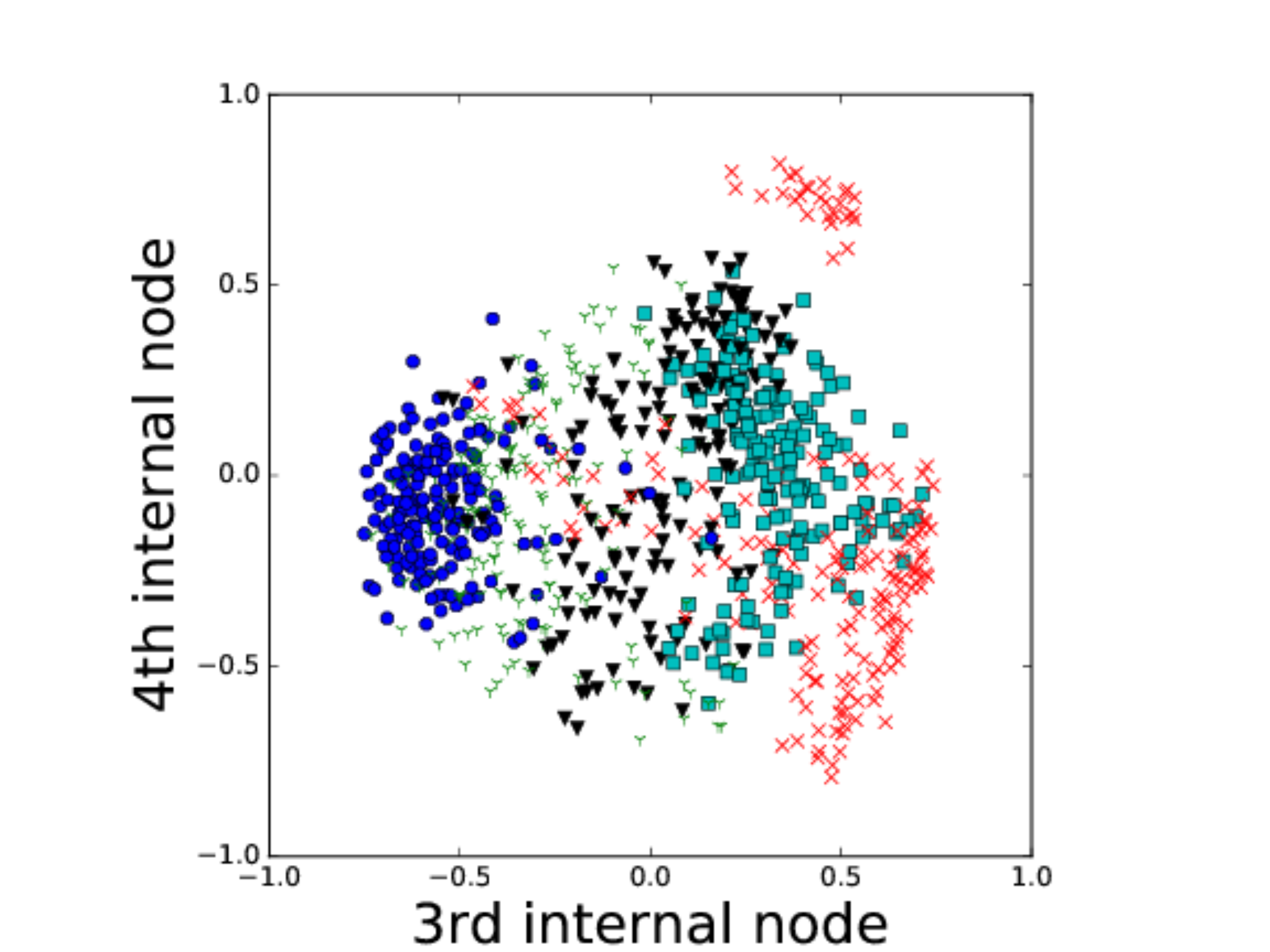}} 
\end{tabular}\\
AE \\
\begin{tabular}{ccc}
\setlength{\tabcolsep}{1.5pt}
\resizebox{0.575\columnwidth}{!}{\includegraphics[trim={2cm 0cm 3cm 0cm},clip]{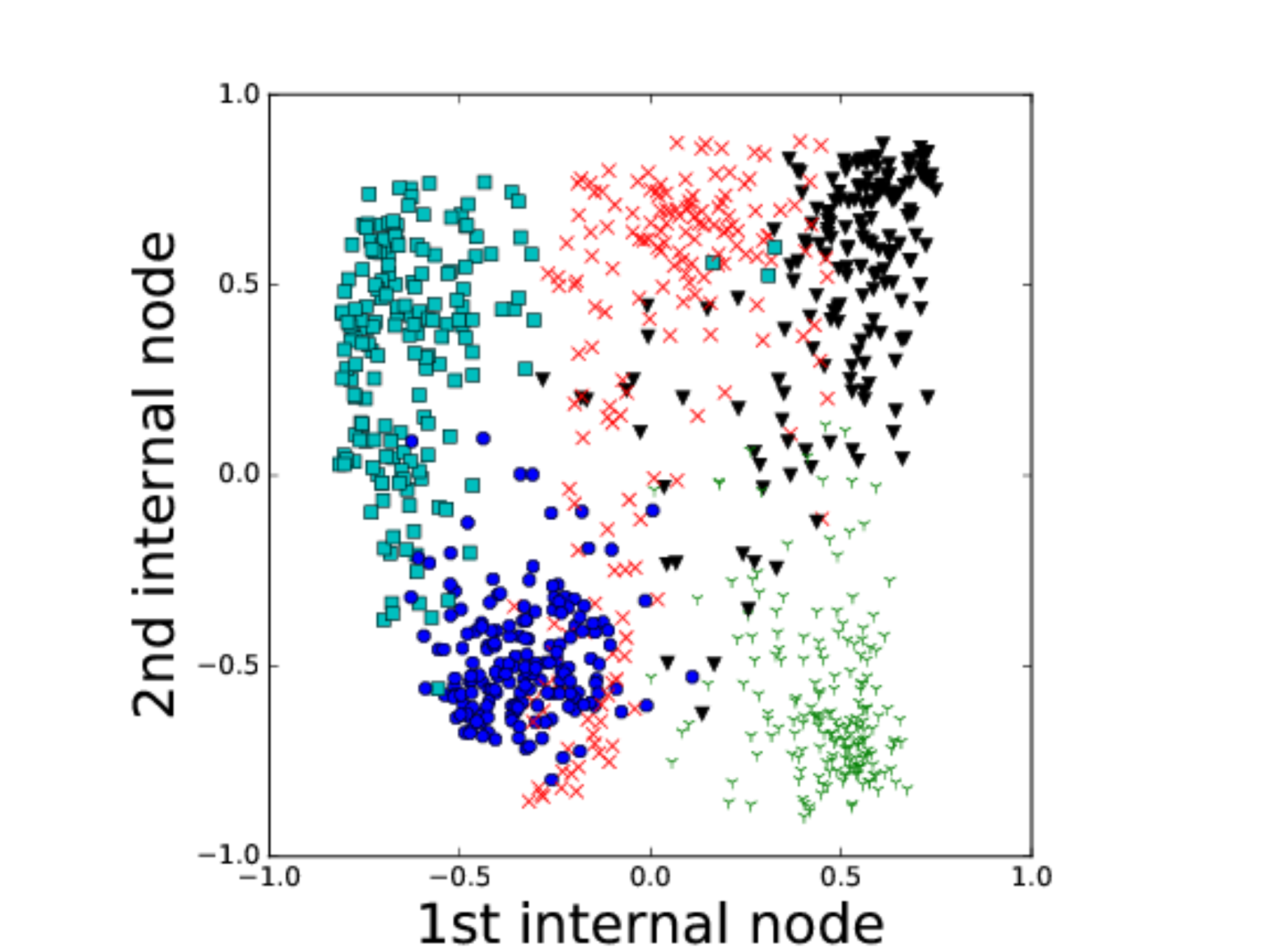}} &
\resizebox{0.575\columnwidth}{!}{\includegraphics[trim={2cm 0cm 3cm 0cm},clip]{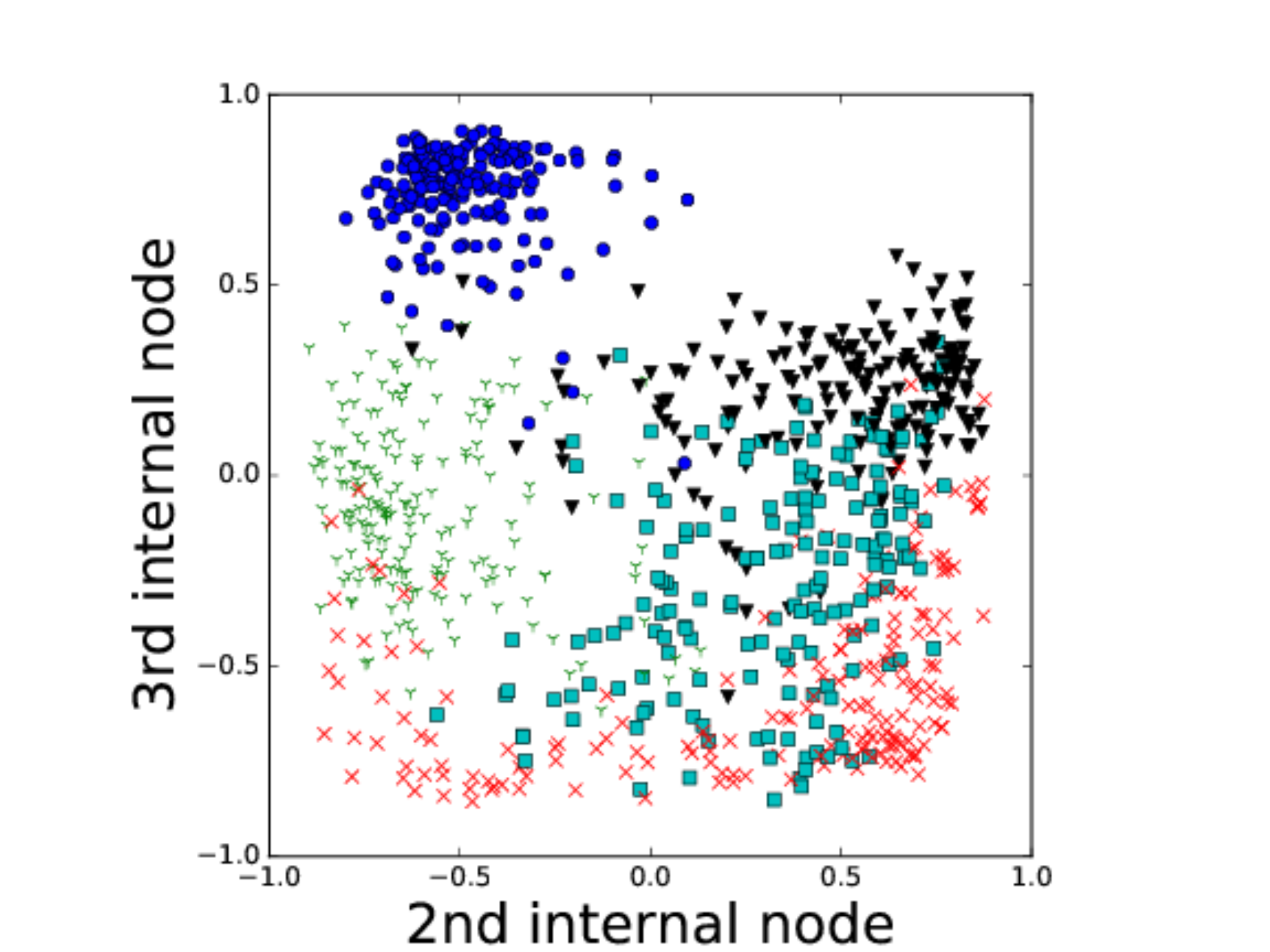}} &
\resizebox{0.575\columnwidth}{!}{\includegraphics[trim={2cm 0cm 3cm 0cm},clip]{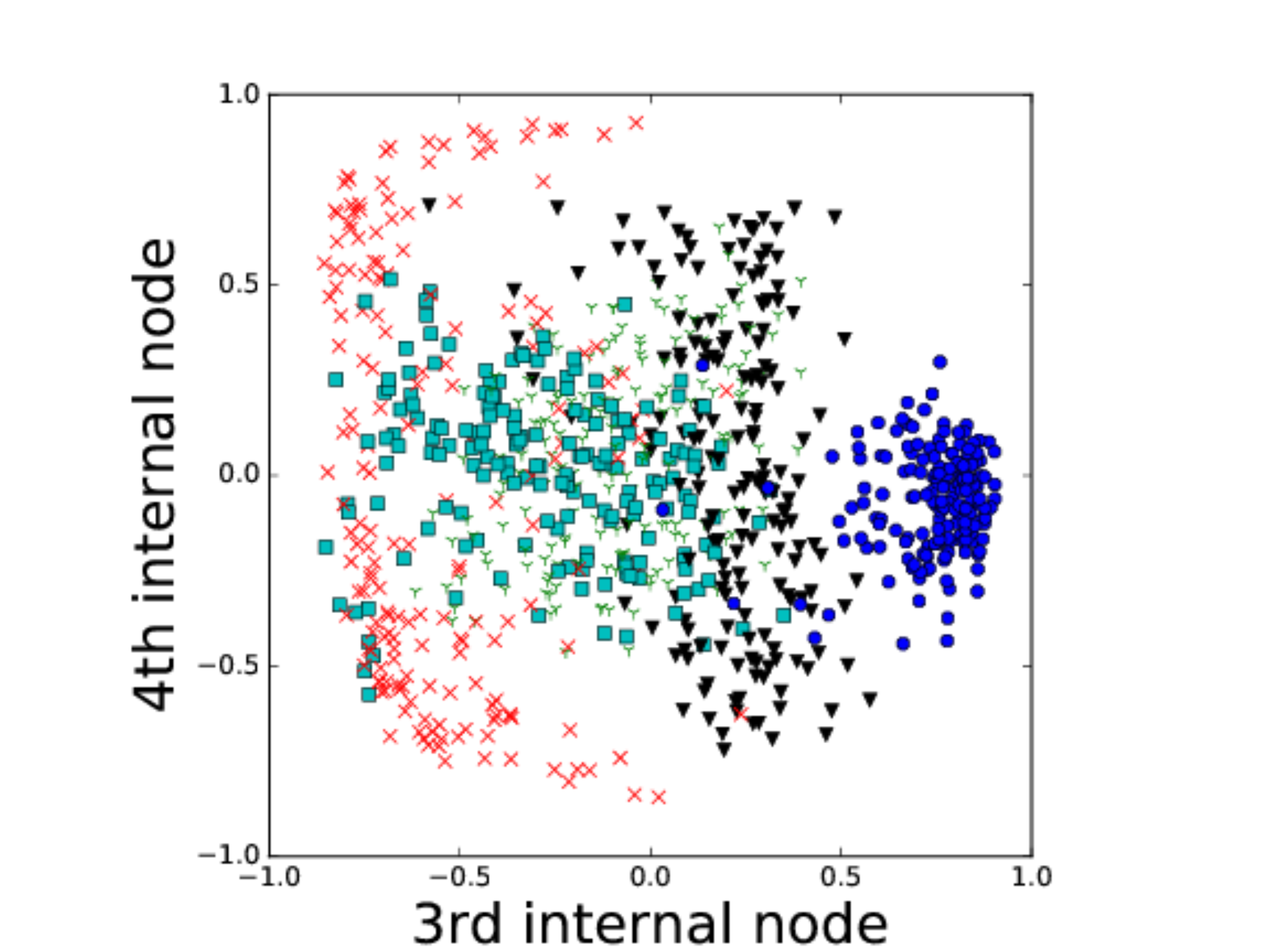}} 
\end{tabular}
\\
GN
\end{tabular}
\end{center}
\caption{Outcomes of four inner nodes using AE and GN algorithm for 5-class handwritten digit data. (The output of the 1st node for both cases is very close to the first principal component (See supplementary information).}
\label{AE_graph}
\vskip -0.2in
\end{figure*}

\subsection{Quality of Features}

First, we appeal to the results on the USPS digits data 
in the Figures~\ref{fig:features-layers} and~\ref{fig:sup-features-layers}
to illustrate the qualitative difference between our greedy 
node-by-node features and the features produced by the standard
layer-by-layer pre-training. 
When you learn all the features in a layer simultaneously, there are
several local minima that do not correspond to the
natural intuitive features that might be present. Because all the
nodes are being trained simultaneously, multiple features can get 
mixed within one node and this mixing can be compensated for in other nodes.
This results in somewhat counter-intuitive features as are obtained in 
Figures~\ref{fig:features-layers} and~\ref{fig:sup-features-layers} for 
the layer-by-layer approach.
The node-by-node approaches do not appear to suffer from this 
problem, as a simple visual inspection of the features implemented
at each node yield recognizable digits. It is clear that our new algorithms
find more interpretable features.

Even though the visual inspection of the features
is quite difference, we may investigate more closely the dimension reduction
from the original pixel-space to the new feature space.
A common approach is to use scatter plots of one feature against another, and
we give such scatter plots in Figure~\ref{AE_graph}, for the
unsupervised setting. For this experiment, we consider a smaller 
handwritten digit data, generated from the scikit-learn package~\cite{sklearn}. 
Readers can refer to the supplementary information for the details on 
the generation procedure. The simplified
handwritten digit data is a 8 by 8 pixel handwritten digits.
It is known that this data admits good classification accuracy using 
the top-2 PCA features. This means that we should see good class separation 
from a scatter plot of the features of one node against another.
(The higher dimensional digits data cannot effectively be visualized in 
2-dimensions.)

In Figure~\ref{AE_graph}, the top row shows scatter plots of one node's features
against another node's features for the standard layer-by-layer algorithm.
Good class separation is obtained even after projecting
to 2-dimensions (there are 5 classes).
The bottom row shows the similar plots for our unsupervised algorithms.
The similarity between the scatter plots indicates that our features
and layer-by-layer features are comparable in terms of classification 
accuracy (we will investigate this in depth in the next section).

\emph{Summary:} Our node-by-node features are 
more interpretable and appear as effective as layer-by-layer
features.


\subsection{Efficiency and Classification Performance} \label{speed}

We have argued theoretically that our algorithms are more efficient 
than layer-by-layer pre-training.
The rationale
is that the training is distributed onto each inner node 
by the partitioning of the data.
We first 
compare the running time in practice  
of supervised and unsupervised pre-training for our algorithms (GN and GCN)
with the standard layer-by-layer algorithms. 
For a concrete comparison, we use a fixed number of
iterations (300) for pre-training with an initial 
learning rate 0.001; we use 500 iterations to train the final 
output layer with logistic regression, with an initial learning rate 0.002;
and finally, we use 
20 iterations of backpropagation to fine-tune all the 
weights with fixed learning rate 0.001. 
We do not use significant learning rate decay.

An $L_2$ regularization term is used with all the algorithms to 
help with overfitting, with a regularization parameter of \math{\lambda=1}. 
It should be noted that this "weight decay" 
regularization term is not always necessary with deep learning since
the flexibility of the network is already significantly diminished. 
In this paper, we are not trying to optimize all hyperparameters 
to get the highest possible test performance for certain data sets.
Instead we try to compare the four algorithm with one 
predefined set of parameters.
Our goal is to see whether there is a deterioration in classification accuracy
due to additional constraints placed by training the deep-network using 
greedy node-by-node learning.
Table~\ref{running-time} shows the results for running time. 
A 256-200-150-10 network structure was used. 
\math{RT} and \math{PT} are the times for the entire algorithm
and for only the pre-training respectively (the entire algorithm
includes fine-tuning and data pre-processing). 
7291 data are in the training set and 2007 are in the test set. 
All the experiment was done using a single Intel i-7 3370 3.4GHz CPU. 
\begin{table}[t] 
\caption{Running-time and performance.
SV (supervised)
and USV (unsupervised) are the standard layer-by-layer algorithms; 
GN (unsupervised) and GCN (supervised) are our algorithms.
\label{running-time}}
\vskip 0.15in
\begin{center}
\begin{small}
\begin{sc}
\begin{tabular}{lllp{1.2cm}p{1.2cm}}
\hline
\abovespace\belowspace
Type & \math{RT} (s) & \math{PT} (s) & Training score & Test score \\
\hline
\abovespace
SV 	      &1363 &1293 &1.000 &0.939\\
USV 	      &4112 &4042 &0.999 &0.933\\
GN 	      &674 &604 &0.999 &0.931\\
\belowspace
GCN 	     &409 &339 &0.998 &0.923\\
\hline    
\end{tabular}
\end{sc}
\end{small}
\end{center}
\vskip -0.1in
\end{table}  
As shown in Table~\ref{running-time}, all algorithms show comparable 
training and test performance, however our algorithms can give an order of 
magnitude speed-up.

\emph{Impact of Amnesia}
The amnesia factor \math{A} is used to train node $i$, 
given the stored results of forward propagation for the previous 
$i-1$ nodes. The stored output is similar to  keeping a ``memory'' 
of previous learned knowledge \emph{within the same representation layer}. 
Node $i$ should learn the new feature with the 
previous information ``in mind''. 
We found that best results are obtained with an amnesia factor between
0 (no memory) and 1.0 (no loss of memory). 
This suggests that some coordination between nodes in a layer is
useful, but too much coordination adds too many constraints.
Our results indicate that amnesia is very important 
for both our new algorithms, and it is not 
a surprise. 
Table~\ref{AF} shows the result of a 256-200-150-10 network with learning 
rate 0.001 and a variety of amnesia factors. 
All the other settings not changed from 
the previous experiment.

\begin{table}[!ht] 
\caption{Effect of the amnesia factor (learning rate: 0.001).
\label{AF}}
\vskip 0.15in
\begin{center}
\begin{small}
\begin{sc}

\begin{tabular}{p{1.4cm}@{\hspace*{0.6cm}}p{1.8cm}@{\hspace*{0.6cm}}p{0.9cm}}
\hline
\abovespace
Amnesia factor& Training score& Test score \\[0.5ex]
\hline
\abovespace
1.0  &0.998 &0.923\\
0.9 &0.999 &0.922 \\
0.8 &0.999 &0.926\\
0.7 &0.999 &0.932\\
0.6 &0.999 &0.927\\
\bf 0.5 &0.999 &\textbf{0.931}\\
\bf 0.4 &0.999 &\textbf{0.934}\\
\belowspace
0.0 &0.974 &0.915 \\
\hline    
\end{tabular}
\end{sc}
\end{small}
\end{center}
\vskip -0.1in
\end{table} 

The results in Table~\ref{AF} suggest that a 
resonable choice for the amnesia factor is \math{A\in[0.4,0.5]},
which is far from both 1 (full coordination) and 0 (no memory that
can result in redundant features).
A non-zero amnesia factor applies a balance between 
two types of representations. A higher AF will lean to 
generative model of inputs but induce numerical issue for training. 
A lower AF will learn a local model of inputs but result in loss 
of information through redundant features. 
The optimal amnesia factor may depend on size 
and depth of the layer. We did not investigate this issue,
keeping \math{A} constant for each layer.
We also did not investigate different ways to distribute the 
data among the nodes. Our simple method works well and is
efficient.

\begin{table*}[t] 
\caption{Comparison between algorithms for multiple data sets showing
training / {\bf test} accuracy.}
\label{data_extra}
\vskip 0.15in
\begin{center}
\begin{small}
\begin{sc}
\begin{tabular}{cccccc}
\hline
\abovespace\belowspace
&&\multicolumn{2}{c}{Layer-by-Layer}&\multicolumn{2}{c}{Greedy Node-by-Node}\\
Data set &Network &SV &USV &GN &GCN \\ 
\hline
\abovespace
MUSK 	&120-80 	&1.000 / \bf 0.991 &1.000 / \bf 0.993 &1.000 / \bf 0.984 &1.000 / \bf 0.989\\
ISOLET 	&260-78 	&0.997 / \bf 0.946 &0.999 / \bf 0.953 &1.000 / \bf 0.944 &1.000 / \bf 0.935\\
CNAE-9	&594-396 &1.000 / \bf 0.954 &0.996	 / \bf 0.944 &0.995 / \bf 0.940 &0.994 / \bf 0.921\\
MADELON	&800-400-200 &1.000 / \bf 0.560 &1.000 / \bf 0.550 &1.000 / \bf 0.645 &1.000 / \bf 0.652\\
CANCER 	&50-40	&0.984 / \bf 0.973 &0.987 / \bf 0.957 &0.979 / \bf 0.968 &0.979 / \bf 0.973\\
BANK	&30-20		&0.949 / \bf 0.886 &0.912 / \bf 0.905 &0.910 / \bf 0.902 &0.913 / \bf 0.896\\
NEWS	&50-40		&0.708 / \bf 0.632 &0.673 / \bf 0.646 &0.689 / \bf 0.646 &0.692 / \bf 0.637\\
POKER	&40-40		&0.718 / \bf 0.579 &0.631 / \bf 0.622 &0.650 / \bf 0.620 &0.629 / \bf 0.624\\
\belowspace
CHESS		&60-40-40		&1.000 / \bf 0.962 &1.000 / \bf 0.867 &0.890 / \bf 0.867 &0.972 / \bf 0.936\\	 
\hline
\end{tabular}
\end{sc}
\end{small}
\end{center}
\vskip -0.1in
\end{table*} 
Finally, we 
give an extensive comparison of the out-of-sample performance between
the four algorithms on several additional 
data sets in Table~\ref{data_extra}.
Nine data sets obtained from UCI machine learning 
repository~\cite{Lichman2013} were used to compare the algorithms. 
Test performance is computed using a validation set which is typically
approximately 
30\% of the data unless the test set is provided by the repository. 
For GCN, it is convenient to set the number of nodes in each hidden layer 
to be divisible by the number of classes. We used the same architecture for 
all algorithms. 
Here are some relevant information for each
data set.
\begin{center}
\begin{sc}
\begin{tabular}{ccccc}\hline
data set&\math{N}&\math{d}&\math{\eta}&\math{A}\\\hline\\[-8pt]
MUSK 	&6598&166&0.01      &0.4 \\
ISOLET 	&6238&617&0.0001&0.4\\
CNAE-9	&1080&856&0.01-0.1&0.4\\
MADELON	&4400&500&0.01&0.4\\
CANCER 	&699&10&0.01&0.4\\
BANK	&4521&16&0.01&0.4\\
NEWS	&39797&60&0.01&0.4\\
POKER	&78211&10&0.01&0.4\\
CHESS     &9149&6&0.01&0.4
\\\hline
\end{tabular}
\end{sc}
\\[3pt]
\parbox{0.85\columnwidth}{\footnotesize 
(\math{N=} \# data points;
\math{d=} dimension (\# attributes);\\
\math{\eta=} initial learning rate of SGD;
\math{A=} amnesia factor.)}
\end{center}
MUSK (version 2) is a molecule classification task with
data on molecules that are judged by experts 
as musks and non-musks. 
ISOLET contains audio information for 26 letters spoken by 
human speakers. 
CNAE-9 contains 1080 documents of text business descriptions for
 Brazilian companies categorized into 9 categories.
CNAE-9 is very sparse, so 
a learning rate up to 0.1 can produce a reasonable performance for 
our new algorithms while the classical algorithms use a learning rate of 0.01.
Better performance results if one specializes the learning rate to the
algorithm 
(for example, GCN with learning rate of 0.15 has the better 
test score of 0.935). 
MADELON is an artificial dataset used in the NIPS 2003 feature selection 
challenge. 
CANCER is the Wisconsin breast cancer data.  
BANK contains bank marketing data. 
NEWS contains news popularity data. 
POKER is part of the 
poker hand data set (class 0 and 1 are removed to get a better class balance).
CHESS is the 
king-rook vs. king data set (KRK) with 4 classes 
(``draw'', ``eight'', ``eleven'', ``fifteen'') 
from the original data set with 6 attributes. 
Table ~\ref{data_extra} shows the comparison between 
the layer-by-layer supervised, unsupervised, GN and GCN algorithms. 

\emph{Summary:}
The performance of our algorithms is comparable
with standard layer-by-layer deep network algorithms, significant speedup and
more interpretable features. The results on additional data sets are
are consistent with the USPS handwriting data set we used throughout 
this paper.

\section{Conclusion and Future Work}

Our goal was to develop a greedy \emph{node-by-node}
deep network algorithm that learns feature representations in each layer
of a deep network sequentially (similar to PCA in the 
linear setting, but different because we partition data among features).
Our two novel deep learning algorithms 
originate from the idea of simulating a human's learning process, namely
building features in a streaming fashion incrementally, using part of the
data to learn each feature.
This is in contrast to classical deep learning algorithms 
which obtain all the hierarchical features in a layer 
for all the objects at the 
same time (train all nodes of a layer simultaneously on all data) --
the human learner learns from one or few objects at a time and is
able to learn new features while leveraging the features it has 
learned on other data. 
Our two new methods, corresponding to supervised learning (GCN) 
and unsupervised learning (GN), do indeed learn one feature of one group 
of training data at a time. Such a design helps to construct more 
human-recognizable features. 
We also developed amnesia, an ability for the
greedy algorithm to control the coordination among the features. 
The results on several datasets reveal that our algorithms
have
a prediction performance comparable with the standard
layer-by-layer  methods
plus the advertised benefits of speed-up and a more interpretable
features.

In the future, we would like to investigate two subproblems. First, 
whether it is possible to further 
exploit the node-by-node paradigm by optimizing
the 
hyper-parameters (learning rate and amnesia)
to obtain superior performance.
Several questions need to be answered here:  
Is there a better way to partition the data? 
How to choose the optimal amnesia factor? 
Should the learning rate be adjusted differently for each inner node or 
each layer? 

Scalability: our algorithms are in some sense
learning in an online fashion, and so  
cannot exploit the matrix-vector and matrix-matrix multiplication approaches
to training that can be easily implemented for multicore or GPU architectures. 
One way to handle such difficulty could be to learn several features at the 
same time in different machines and exchange information 
(as memorization constraints) every few epochs 
(to simulate a group of human learners). 
Distributing our algorithm over different models of parallel
computing appears to be a challenging problem.
The adaption of the algorithm to parallel computing will surely 
require creative upgrades in the algorithm design. 